\documentclass[referee,sn-mathphys,Numbered,pdflatex]{sn-jnl}

\usepackage{graphicx}%
\usepackage{multirow}%
\usepackage{amsmath,amssymb,amsfonts}%
\usepackage{amsthm}%
\usepackage{mathrsfs}%
\usepackage[title]{appendix}%
\usepackage{textcomp}%
\usepackage{manyfoot}%
\usepackage{booktabs}%
\usepackage{algorithm}%
\usepackage{algorithmicx}%
\usepackage{algpseudocode}
\usepackage{listings}%
\usepackage{pifont}

\usepackage{hyperref}
\usepackage{theoremref}
\usepackage{mathdots}
\usepackage[table]{xcolor}
\usepackage{diagbox}
\usepackage{glossaries}
\newcommand{\xmark}{\ding{55}}

\theoremstyle{thmstyleone}%
\newtheorem{theorem}{Theorem}
\newtheorem{corollary}{Corollary}

\theoremstyle{thmstyletwo}%

\theoremstyle{thmstylethree}%

\makeglossaries
\newglossaryentry{sp}{
    name=SP,
    first={Statistical Parity (SP)},
    description={Statistical Parity; a common fairness definition requiring the equality of positive outcomes across groups}
}
\newglossaryentry{ai}{
    name={AI},
    first={Artificial Intelligence (AI)},
    description={Artificial Intelligence; the scientific field concerned with developing theory and systems that perform tasks requiring intelligence}
}
\newglossaryentry{aaspe}{
    name={AASPE},
    first={Average Absolute Statistical Parity Error (AASPE)},
    description={Average Absolute Statistical Parity Error; the error measure used to estimate the performance of the method introduced in this work}
}
\newglossaryentry{xai}{
    name={XAI},
    first={Explainable Artificial Intelligence (XAI)},
    description={Explainable Artificial Intelligence; the subfield of Artificial Intelligence concerned with making systems that can be interpreted and explained}
}
\newglossaryentry{ml}{
    name={ML},
    first={Machine Learning (ML)},
    description={Machine Learning; the subfield of Artificial Intelligence concerned with building systems that can learn a task without following explicit instructions}
}
\newglossaryentry{eqodds}{
    name={EOdd},
    first={Equalized Odds (EOdd)},
    description={Equalized Odds; a common fairness definition requiring equality of false positive and true positive rates across groups}
}
\newglossaryentry{eqopp}{
    name={EOpp},
    first={Equality of Opportunity (EOpp)},
    description={Equality of Opportunity; a common fairness definition requiring equality of true positive rates across groups}
}
\newglossaryentry{preq}{
    name={PrEq},
    first={Predictive Equality (PrEq)},
    description={Predictive Equality; a common fairness definition requiring equality of false positive rates across groups}
}
\newglossaryentry{dt}{
    name={DT},
    first = {Decision Tree (DT)},
    description={Decision Tree; An interpretable, rule-based machine learning model that is used for classification tasks},
    plural={DTs},
    firstplural={Decision Trees (DTs)}
}
\newglossaryentry{dp}{
    name={DP},
    first={Differential Privacy (DP)},
    description={Differential Privacy; A class of methods to share information in a privacy aware manner}
}
\newglossaryentry{uar}{
    name={UAR}, 
    first = {UAR (Unweighted Average Recall)}, 
    description={Unweighted Average Recall; A performance metric used for classification tasks that measures the recall for each class}
}
\newglossaryentry{pafer}{
    name={PAFER},
    first={Privacy-Aware Fairness Estimation of Rules (PAFER)},
    description={Privacy-Aware Fairness Estimation of Rules; The proposed and studied method in this work that can estimate fairness while respecting privacy}
}

\begin{document}

\title[Privacy Constrained Fairness Estimation for Decision Trees]{Privacy Constrained Fairness Estimation for Decision Trees}


\author*[1]{\fnm{Florian} \spfx{van der} \sur{Steen}}\email{florianvandersteen@gmail.com}

\author[2]{\fnm{Fr\'e} \sur{Vink}}\email{f.t.vink@minfin.nl}

\author[1]{\fnm{Heysem} \sur{Kaya}}\email{h.kaya@uu.nl}

\affil*[1]{\orgdiv{Department of Information and Computing Sciences}, \orgname{Utrecht University}, \orgaddress{\street{Princetonplein 5}, \city{Utrecht}, \postcode{3584 CC},  \country{the Netherlands}}}

\affil[2]{\orgdiv{Responsible AI Team}, \orgname{Dutch Central Government Audit Service}, \orgaddress{\street{Korte Voorhout 7}, \city{Den Haag}, \postcode{2511 CW}, \country{the Netherlands}}}



\abstract{The protection of sensitive data becomes more vital, as data increases in value and potency. Furthermore, the pressure increases from regulators and society on model developers to make their \gls{ai} models non-discriminatory. To boot, there is a need for interpretable, transparent \gls{ai} models for high-stakes tasks. In general, measuring the fairness of any \gls{ai} model requires the sensitive attributes of the individuals in the dataset, thus raising privacy concerns. In this work, the trade-offs between fairness, privacy and interpretability are further explored. We specifically examine the \gls{sp} of \glspl{dt} with \gls{dp}, that are each popular methods in their respective subfield. We propose a novel method, dubbed Privacy-Aware Fairness Estimation of Rules (PAFER), that can estimate \gls{sp} in a \gls{dp}-aware manner for \glspl{dt}. \gls{dp}, making use of a third-party legal entity that securely holds this sensitive data, guarantees privacy by adding noise to the sensitive data. We experimentally compare several \gls{dp} mechanisms. We show that using the Laplacian mechanism, the method is able to estimate \gls{sp} with low error while guaranteeing the privacy of the individuals in the dataset with high certainty. We further show experimentally and theoretically that the method performs better for \glspl{dt} that humans generally find easier to interpret. }

\keywords{responsible AI, fairness, interpretability, differential privacy}



\maketitle

\section{Introduction}\label{sec1}

The methods from the scientific field of \Gls{ai}, and in particular \Gls{ml}, are increasingly applied to tasks in socially sensitive domains. Due to their predictive power, \Gls{ml} models are used within banks for credit risk assessment~\cite{zhu_2019_study}, aid decisions within universities for new student admissions~\cite{bickel_1975_sex} and aid bail decision-making within courts~\cite{chouldechova_fair_2017}. Algorithmic decisions in these settings can have fargoing impacts, potentially increasing disparities within society. Numerous notorious examples exist of algorithms causing harm in this regard. In 2015, Google Photos new image recognition model classified some black individuals as gorillas \cite{barr_google_2015}. This led to the removal of the category within Google Photos. A report by Amnesty International concluded that the Dutch Tax \& Customs administration used a model for fraud prediction that discriminated against people with multiple nationalities \cite{noauthor_xenophobic_2021}.

The application of \gls{ml} should clearly be done responsibly, giving rise to a field that considers the fairness of algorithmic decisions. Fair \Gls{ml} is a field within \Gls{ai} concerned with assessing and developing fair \gls{ml} models. Fairness in this sense closely relates to equality between groups and individuals. The main notion within the field is that models should not be biased, that is, have tendencies to over/underperform for certain (groups of) individuals. This notion of bias is different from the canonical definition of bias in statistics, i.e.\ the difference between an estimator's expected value and the true value. Essentially, similar individuals should be treated similarly, and decisions should not lead to unjust discrimination. Non-discrimination laws for \Gls{ai} exist within the EU \cite{gdpr_2016} and more are upcoming \cite{aiact_2021}. The Dutch government now has a register of all the algorithms used within it \cite{noauthor_algoritmeregister_2022}.

An additional property that responsible \gls{ml} models should have, is that they are interpretable. Models of which the decision can be explained, are preferred as they aid decision-making processes affecting real people. In a loan application setting, users have the right to know how a decision came about \cite{mortgage_2014}. The field of \gls{xai}, is concerned with building models that are interpretable and explainable. 

Inherently, \Gls{ml} models use data. Thus, there is also a tension between the use of these models and privacy, especially for socially sensitive tasks. Individuals have several rights when it comes to data storage, such as the right to be removed from a database \cite{gdpr_2016}. It is also beneficial for entities to guarantee privacy so that more individuals trust the entity with their data. Some data storage practices are discouraged such as the collection of several protected attributes \cite{gdpr_2016}. These attributes, and thus the storage practices thereof, are sensitive. Examples include the religion, marital status, and gender of individuals. In industrial settings, numerous data leaks have occurred. Social media platforms are especially notorious for privacy violations, with Facebook even incurring data breaches on multiple occasions \cite{cadwalladr_revealed_2018, noauthor_losing_2019}. The report by Amnesty International also concluded that the Dutch Tax \& Customs Administration in the Dutch childcare benefits scandal failed to safely handle the sensitive private data of thousands of individuals, while they used a biased model \cite{noauthor_xenophobic_2021}. This work will investigate these three pillars of Responsible \gls{ai}, investigating a novel method that is at the intersection of these three themes. 

To assess and improve fairness precisely, one needs the sensitive attributes of the individuals that a \Gls{ml} model was trained on. But these are often absent or have limited availability, due to privacy considerations. Exactly here lies the focal point of this work, the assessment of the fairness of \Gls{ml} models, while respecting the privacy of the individuals in the dataset. These antagonistic goals make for a novel, highly constrained, and hence difficult problem. A focus is placed on \glspl{dt}, a class of interpretable models from \gls{xai} since these types of models are likely to be used in critical tasks involving humans due to the GDPR (in particular Art. 22)~\cite{gdpr_2016} and its national implementations. There are thus four goals we try to optimize in this work: fairness, privacy, interpretability, and predictive performance.  

\subsection{Research Questions}\label{subsec:rqs}
The main goal of this work is to develop a method that can estimate the fairness of an interpretable model with a high accuracy while respecting privacy. A method, named \gls{pafer}, is proposed that can estimate the fairness of a class of interpretable models, \glspl{dt}, while respecting privacy. The method is thus at the intersection of these three responsible \gls{ai} pillars. The research questions (RQs), along with their research subquestions, (RSQs) are:

\begin{enumerate}
    \item[] \textbf{RQ1} What is the optimal privacy mechanism that preserves privacy and minimizes average Statistical Parity error?
    \begin{enumerate}
        \item []\textbf{RSQ1.1} Is there a statistically significant mean difference in Absolute Statistical Parity error between the Laplacian mechanism and the Exponential mechanism?
    \end{enumerate}
    \item[] \textbf{RQ2} Is there a statistically significant difference between the Statistical Parity errors of \gls{pafer} compared to other benchmarks for varying Decision Tree hyperparameter values?
    \begin{enumerate}
        \item[] \textbf{RSQ2.1} At what fractional \texttt{minleaf} value is \gls{pafer} significantly better at estimating Statistical Parity than a random baseline?
    \end{enumerate}
\end{enumerate}

\subsection{Outline}
The remainder of the paper is organized as follows. The upcoming \autoref{sec:prelims} will provide the theoretical background, which is followed by \autoref{sec:relatedwork} that covers the related literature. \hyperref[sec:proposedmethod]{Section~\ref*{sec:proposedmethod}} describes the novel method that is proposed in this work. Subsequently, \autoref{sec:evaluation} describes the performed experiments, their results, and thorough analysis. Finally, \autoref{sec:conclusionfutwork} concludes with limitations and future directions. 

\section{Preliminaries}\label{sec:prelims}
This section discusses work related to the research objectives and provides background to the performed research. \hyperref[subsec:fairnessmetrics]{Subsection~\ref*{subsec:fairnessmetrics}} describes fairness theory, \autoref{subsec:interpretmodels} provides background on interpretable models and \autoref{subsec:privadef} explains notions of privacy. 

\subsection{Fairness Definitions}\label{subsec:fairnessmetrics}
Fairness in an algorithmic setting relates to the way an algorithm handles different (groups of) individuals. Unjust discrimination\footnote{What exactly is \textbf{unjust} discrimination is a social construct and changes over time \cite{berendt_2014_better}.} is often the subject when examining the behavior of algorithms with respect to groups of individuals. For this work, only fairness definitions relating to supervised \Gls{ml} were studied, as this is the largest research area within algorithmic fairness. 

In 2016, the number of papers related to fairness surged. Partly, due to the new regulations such as the European GDPR \cite{gdpr_2016} and partly due to a popular article by ProPublica which examined racial disparities in recidivism prediction software \cite{mattu_machine_2016}. Because of the young age of the field and the sudden rise in activity, numerous definitions of fairness have been proposed since. Most of the definitions also simultaneously hold multiple names; this section aims to include as many of the names for each definition. 

The performance-oriented nature of the \Gls{ml} research field accelerated the development of fairness metrics, quantifying the fairness for a particular model. The majority of the definitions can therefore also be seen, or rewritten, as a measuring stick for the fairness of a supervised \Gls{ml} model. This measurement may be on a scale, which is the case for most group fairness definitions, or binary, which is the case for some causal fairness definitions. 

The fairness definitions, namely the mathematical measures of fairness, can be categorized into group fairness, individual fairness and causal fairness \cite{verma_fairness_2018}. Considering the space limitations and the relevance to our work, in this section, we will focus on group fairness and provide the definitions of the most prominent measures used in the literature. Group fairness is the most popular type of fairness definition as it relates most closely to unjust discrimination. Individuals are grouped based on a sensitive, or protected attribute, \(A\), which partitions the population. This partition is often binary, for instance when \(A\) denotes a privileged and unprivileged group. In this subsection, we assume a binary partition for ease of notation, but all mentioned definitions can be applied to $\mathcal{K}$-order partitions. Some attributes are protected by law, for example, gender, ethnicity and age.

The setting for these definitions is often the binary classification setting where \(Y \in \{0, 1\}\), with \(Y\) as the outcome. This is partly due to ease of notation, but more importantly, the binary classification setting is common in impactful prediction tasks. Examples of impactful prediction tasks are granting or not granting a loan \cite{zhu_2019_study}, accepting or not accepting students to a university \cite{bickel_1975_sex} and predicting recidivism after a certain period \cite{chouldechova_fair_2017}. In each setting, a clear favorable (1) and unfavorable (0) outcome can be identified. Thus, we assume the binary classification setting in the following definitions. 

\subsubsection{Statistical Parity}
Statistical Parity (\gls{sp}) is a decision-based definition, which compares the different positive prediction rates for each group \cite{dwork_fairness_2012}. \Gls{sp}, also known as demographic parity, equal acceptance rate, total variation or the independence criterion, is by far the most popular fairness definition. The mathematical definition is:
\begin{equation}\label{eq:statpar}
    {\text{SP}} = p(\hat{Y}=1|A=1) - p(\hat{Y}=1|A=0),
\end{equation}
where \(\hat{Y}\) is the decision of the classifier. An example of \Gls{sp} would be the comparison of the acceptance rates of males and females to a university. 

Note that \autoref{eq:statpar} is the \Gls{sp}-difference but the \Gls{sp}-ratio also exists. US law adopts this definition of \gls{sp} as the 80\%-rule \cite{eighty_rule_1979}. The 80\%-rule states that the ratio of the acceptance rates must not be smaller than 0.8, i.e.\ 80\%. Formally:
\begin{equation}
    {\text{80\%-rule}} \; = 0.8 \leq \frac{p(\hat{Y}=1|A=1)}{p(\hat{Y}=1|A=0)} \leq 1.25,
\end{equation}
where the fraction is the \Gls{sp}-ratio. \Gls{sp} is easy to compute and merely uses the model's predictions. \gls{sp} therefore does not require labelled data. These advantages make it one of the most used fairness definitions. 

\subsubsection{Equalized Odds}\label{subsubsec:eqodds}
Another, also very common, fairness definition is the \Gls{eqodds} metric \cite{hardt_equality_2016}. It is also known as disparate mistreatment or the separation criterion. \Gls{eqodds} requires that the probabilities of being correctly positively classified and the probabilities of being incorrectly positively classified are equal across groups. Thus, the definition is twofold; both false positive classification probability and true positive classification probability should be equal across groups. Formally:
\begin{equation}
    {\text{EOdd}} = p(\hat{Y}=1|Y=y,A=1) - p(\hat{Y}=1|Y=y,A=0), \; \; y \in \{0, 1\}.
\end{equation}
An advantage of \Gls{eqodds} is that, unlike \Gls{sp}, when the predictor is perfect, i.e.\ \(Y = \hat{Y}\), it satisfies \Gls{eqodds}. 

\subsubsection{Equality of Opportunity}
A relaxation of \Gls{eqodds} is the fairness definition \Gls{eqopp} \cite{hardt_equality_2016}. It just requires the equality of the probabilities of correctly predicting the positive class across groups. In other words, where \Gls{eqodds} requires that both true positive and false positive classification rates are equal across groups, \Gls{eqopp} only requires the former. Formally:
\begin{equation}
    {\text{EOpp} = p(\hat{Y}=1|Y=1,A=1) - p(\hat{Y}=1|Y=1,A=0)}.
\end{equation}
An advantage of \Gls{eqopp} is that it is not a bi-objective, and thus is more easily optimized for compared to \Gls{eqodds}.

\subsection{Interpretable Models}\label{subsec:interpretmodels}
This subsection outlines a class of models with inherently high interpretability, \glspl{dt}, that are central to this work. The interpretability of a model is the degree to which the classifications and the decision-making mechanism can be interpreted. The field of \Gls{xai} is concerned with building systems that can be interpreted and explained. Complex systems might need an explanation function that generates explanations for the outputs of the system. Some methods may inherently be highly interpretable, requiring no explanation method, such as \glspl{dt}. Interpretability may be desired to ensure safety, gain insight, enable auditing or manage expectations. 

\subsubsection{Decision Trees (DTs)} 
A \gls{dt} is a type of rule-based system that can be used for classification problems. The structure of the tree is learned from a labelled dataset. \glspl{dt} consist of nodes, namely branching nodes and leaf nodes. The upper branching node is the root node. To classify an instance, one starts at the root node and follows the rules which apply to the instance from branching node to branching node until no more rules can be applied. Then, one reaches a decision node, also called a leaf node. Every node holds the instances that could reach that node. Thus, the root node holds every instance. Decision nodes classify instances based on the class that represents the most individuals within that node. 

There are two effective ways to determine the structure of a \gls{dt}, given a labelled dataset. The most common way is to have a function that indicates what should be the splitting criterion in each branching node. These heuristic functions look at splitting criteria to partition the data in the node such that each partition is as homogeneous as possible w.r.t.\ class. An example of such a heuristic is entropy, intuitively defined as the degree to which the class distribution is random in a partition. A greedy process then constructs the tree, picking the best split in each individual node. Optimal \glspl{dt} are a newer set of approaches, that utilize methods from dynamic programming and constrained optimization \cite{bertsimas2017optimal}. Their performance is generally better as they approach the true \gls{dt} more closely than greedily constructed \glspl{dt}. However, their construction is computationally heavy. 

The interpretability of a \gls{dt} is determined by several factors. The main factor is its height, the number of times the \gls{dt} partitions the data. Very shallow Decision Trees are sometimes called Decision Stumps \cite{oliver1994averaging}. The \texttt{minleaf} \gls{dt} hyperparameter also influences the interpretability of a \gls{dt}. The \texttt{minleaf} value constrains how many instances should minimally hold in a leaf node. The smaller the value, the more splits are required to reach the set \texttt{minleaf} value. Optimal \glspl{dt} cannot have a tall height due to their high computational cost. Greedy \glspl{dt} can be terminated early in the construction process to maintain interpretability. Closely related to height is the number of decision nodes in the tree. This also influences the interpretability of \glspl{dt}, as the more decision nodes a \gls{dt} has, the more complex the \gls{dt} is. Finally, \glspl{dt} built with numeric features might become uninterpretable because they use the same numeric feature over and over, leading to unintuitive decision boundaries. 

In general, \glspl{dt} are interpretable because they offer visualizations and use rules, which are both easy to understand for humans \cite{molnar2022}. Major disadvantages of \glspl{dt} include their incapability of efficiently modeling linear relationships and their sensitivity to changes in the data. Still, their performance, especially ensembles of \glspl{dt}, are state-of-the-art for prediction tasks on tabular data \cite{borisov2022deep}.

\subsection{Privacy Definitions}\label{subsec:privadef}
The final main pillar of responsible \gls{ai} that this work discusses is privacy. Privacy, in general, is a term that can be used in multiple contexts. In its literal sense, privacy relates to one's ability to make personal and intimate decisions with nothing interfering. In this work, however, privacy refers to the degree of control one has over others accessing personal data about themselves. This is also known as informational privacy. The less personal data others access about an individual, the more privacy the individual has. This subsection discusses several techniques to increase informational privacy. 

\subsubsection{Differential Privacy (DP)}
Differential Privacy (DP) \cite{dwork_differential_2006} is a notion that gives mathematical guarantees on the membership of individuals in a dataset. In principle, it is a promise to any individual in a dataset, namely: `You will not be affected, adversely or otherwise, by allowing your data to be used in any analysis of the data, no matter what other analyses, datasets, or information sources are available' \cite{dwork_algorithmic_2013}. More specifically, an adversary cannot infer if an individual is in the dataset. \gls{dp} can be applied when sharing data, or an analysis of the data. \gls{ml} models are ways of analysing data and therefore can also promise to adhere to \gls{dp}. Another guarantee that \gls{dp} makes is that it is immune to post-processing, i.e.\ \gls{dp} cannot be undone \cite{dwork_algorithmic_2013}. 

\paragraph{Definition}
The promise of \gls{dp} can be mathematically guaranteed up to a parameter \(\varepsilon\). A higher \(\varepsilon\) guarantees more privacy. This parameter \(\varepsilon\) is the privacy budget. The main means of guaranteeing the promise of \gls{dp} is by perturbing the data, i.e.\ adding noise to the data. In the context of building \gls{ml} models, this noise may be added to the parameters of the \gls{ml} model or to its training data. At any rate, there is a query, \(q(\cdot)\), for data\footnote{This query may come from a user of a \gls{ml} model or from a developer that requires training data.}, to which \gls{dp} adds noise. Because \gls{dp} is based on membership inference, the formal definition compares two neighboring datasets, \(D\) and \(D'\), in which only one instance differs. For these datasets, \((\varepsilon, \delta)\)-\gls{dp} formally is:
\begin{equation}
    p(\mathcal{A}(q(D)) \subseteq range(\mathcal{A})) \leq \exp(\varepsilon) \cdot p(\mathcal{A}(q(D')) + \delta \subseteq range(\mathcal{A})),
\end{equation}
where \(\mathcal{A}\) is a randomized mechanism around a query \(q(\cdot)\) and \(range(\mathcal{A})\) is the range of all outcomes the mechanism can have. If $\delta = 0$, $\varepsilon$-\gls{dp} is satisfied. \gls{dp}-mechanisms thus randomize query answers in some way. 

\paragraph{Global Sensitivity}
How much noise ought to be added, depends on the difference the inclusion of one worst-case individual in the dataset makes for the query answer. This is known as the sensitivity, \(\Delta q\), how sensitive a query answer is to a change in the data \cite{dwork_differential_2006}. Formally:
\begin{equation}\label{eq:globalsens}
    \Delta q = \max_{D, D'} \; \lvert \lvert q(D) - q(D') \rvert \rvert _1 ,
\end{equation}
which is also know as the \(\ell_1\)-sensitivity or the global sensitivity. 

\paragraph{Laplace Mechanism}
Several techniques exist to randomize query answers, of which the most common one is the Laplacian mechanism \cite{dwork_differential_2006}, for queries requesting real numbers\footnote{An example of such a query might be: `What is the average age of females in the dataset?'.}. The mechanism involves adding noise to a query answer, sampled from the Laplace distribution, centered at 0 and with a scale equal to \(\frac{\Delta q}{\varepsilon}\). The Laplace mechanism can be formalised as:
\begin{equation}
    \mathcal{A}(D, q(\cdot), \varepsilon) = q(D) + Lap(\frac{\Delta q}{\varepsilon}), 
\end{equation}
where \(Lap(\frac{\Delta q}{\varepsilon})\) is the added Laplacian noise. 

\paragraph{Exponential Mechanism}
A different noise schema is the Exponential mechanism \cite{mcsherry2007mechanism}, used for categorical, utility-related queries\footnote{An example of such a query might be: `What is the optimal attribute to partition the dataset in terms of class?' Such a query can be found in the next subsection.}. For these sorts of queries, a small amount of noise may completely destroy the utility of the query answer. A utility function, \(u_D(r)\), is defined over the categories, \(r \in \mathcal{R}\), for a certain dataset $D$. The exponential mechanism is sensitive w.r.t.\ the utility function, \(\Delta u\), not with respect to changes in \(r\). 
The exponential mechanism can be formally defined as:
\begin{equation}
    p(\mathcal{A}(D, u, \mathcal{R}, \varepsilon) = r) \propto \exp(\frac{\varepsilon u_D(r)}{2 \Delta u}).
\end{equation}
In other words, the probability of the best category being chosen is proportional to \(e^{\frac{\varepsilon u_D(r)}{2 \Delta u}}\).

\paragraph{Gaussian Mechanism}
The Gaussian mechanism adds noise based on the Gaussian distribution, with $\mathcal{N}(0, \sigma)$. The mechanism is similar to the Laplacian mechanism in this sense. \gls{dp} holds if $\sigma \geq \sqrt{2 \ln(\frac{1.25}{\delta})}\frac{\Delta_2}{\varepsilon}$ \cite{dwork_algorithmic_2013}. The term $\Delta_2$ is the global $\ell_2$-sensitivity; instead of using the $\ell_1$-norm in \autoref{eq:globalsens}, $\Delta_2$ uses the $\ell_2$-norm. The Gaussian mechanism can be deemed a more `natural' type of noise, as it adds noise that is often assumed to be present in measurements. A disadvantage is that both $\delta$ and $\varepsilon$ must be in $(0, 1)$, so $\varepsilon$-\gls{dp} can never be met.

\section{Related Work}\label{sec:relatedwork}
This section discusses work related to the research objectives. Whereas the previous section discussed background related to only one pillar of Responsible \gls{ai}, this section will highlight methods at the intersection of these fields. It concludes by relating the proposed method, \gls{pafer}, to the current landscape of methods. 

\subsection{Fair Decision Trees}
Some of the earliest work regarding fair \glspl{dt} was performed by 
Kamiran \& Calders and is now known as Discrimination Aware Decision Trees (DADT). They proposed a Heuristic-Based \gls{dt} that incorporates the homogeneity of the sensitive attribute into the splitting criterion \cite{kamiran_discrimination_2010}. DADT also performs some post-processing s.t. certain decision nodes change their decision. This step is phrased as a KNAPSACK problem \cite{ausiello2012complexity}, and is also solved greedily. 

In terms of optimal \glspl{dt}, Linden et al. achieve excellent results with a method named DPFair \cite{linden_fair_2022}. Their work significantly improves the speed of the work of Jo et al., who formulate the optimal \gls{dt} problem with an additional fairness objective \cite{jo_learning_2022}. 

\subsection{Privacy-aware Decision Trees}
\glspl{dt} with privacy guarantees are best represented by the work of Mohammed et al.. The method, named Private Decision tree Algorithm (PDA), uses the Exponential mechanism and queries the required quantities for greedily building op the \gls{dt}~\cite{mohammed_secure_2015}. For an in-depth overview of \glspl{dt} with privacy guarantees, the reader is referred to \cite{fletcher_decision_2020}. 

\subsection{Fair Privacy-aware models}
There is an upcoming field within responsible \gls{ai} that is aimed at improving fairness, without accessing sensitive data. Prominent examples include Adversarially Reweighted Learning (ARL) and Fair Related Features 
(FairRF) \cite{lahoti_fairness_2020, zhao_towards_2022}, respectively. While we highly value this line of work, it does not allow for the evaluation or estimation of fairness, as the field assumes sensitive attributes are entirely unavailable. Therefore, we consider these methods to be insufficient for our purpose, as we aim to provide guarantees on the degree of fairness a model exhibits, e.g. adherence to the 80\%-rule. 

The method most closely related to ours is named AttributeConceal and was introduced by Hamman et al.. They explore the idea of querying the group fairness metrics \cite{hamman_can_2022}. The scenario they assume is that \gls{ml} developers have some dataset without sensitive attributes for which they build models, and therefore query \gls{sp} and \gls{eqodds} from a data curator. They establish that if the developers have bad intentions, they can identify a sensitive attribute of an individual using one unrealistic query, or two realistic ones. The main idea is that the models, for which they query fairness metrics, differ only on one individual, giving away their sensitive attribute via the answer. This result is then extended using any number of individuals. When the sizes of the groups differ greatly, i.e.\ \(\lvert D_{A=0} \rvert \ll \lvert D_{A=1} \rvert\), using compressed sensing \cite{candes2008introduction}, the number of queries is in \(O(\lvert D_{A=0} \rvert\log(\frac{N}{\lvert D_{A=1} \rvert}))\), with \(N = \lvert D_{A=1} + D_{A=0} \rvert\), the total number of instances. The authors propose a mitigation strategy named AttributeConceal, using smooth sensitivity. This is a sensitivity notion that is based on the worst-case individual in the dataset. \gls{dp} is ensured for any number of queries by adding noise to each query answer. It is experimentally verified that using AttributeConceal, an adversary can predict sensitive attributes merely as well as a random estimator.

\subsection{PAFER \& Related Work}
\autoref{tab:relatedmethods} shows methods from the domain of responsible \gls{ai} that have similar goals to \gls{pafer}. In general, we see a lack of fair, privacy-preserving methods for rule-based methods, specifically \glspl{dt}. Hamman et al.\ investigate the fairness of models in general without giving in on privacy \cite{hamman_can_2022}, but the method lacks validity. The developers, in their setting, do not gain intuition on what should be changed about their model to improve fairness. One class of models that lends itself well to this would be \glspl{dt}, as these are modular and can be pruned, i.e.\ rules can be removed. \glspl{dt} are the state-of-the-art for tabular data \cite{borisov2022deep} and sensitive tasks are often prediction tasks for tabular data\footnote{Examples are university acceptance \cite{bickel_1975_sex}, bail decision making \cite{chouldechova_fair_2017} and credit risk assessment \cite{zhu_2019_study}. }. A method that can identify unfairness in a privacy-aware manner for \glspl{dt} would be interpretable, fair and differentially private, respecting some of the most important pillars of responsible \gls{ai}. \gls{pafer} aims to fill this gap, querying the individual rules in a \gls{dt}. The next section will introduce the method.

\begin{table}[t]
    \caption{Overview of methods that are similar to \gls{pafer}. Fairness-Aware methods are methods that aim to improve or estimate fairness.}
    \begin{tabular}{lccc} \toprule
    Method & Interpretable & Privacy-aware & Fairness-Aware \\
    \midrule 
    DADT \cite{kamiran_discrimination_2010}  & \checkmark & \xmark & \checkmark \\
    DPFair \cite{linden_fair_2022} & \checkmark & \xmark & \checkmark \\ 
    PDA \cite{mohammed_secure_2015} & \checkmark & \checkmark & \xmark \\
    ARL \cite{lahoti_fairness_2020} & \xmark & \checkmark & \checkmark \\
    FairRF \cite{zhao_towards_2022} & \xmark & \checkmark & \checkmark \\
    AttributeConceal \cite{hamman_can_2022} & \xmark & \checkmark & \checkmark \\
    \gls{pafer} & \checkmark & \checkmark & \checkmark \\  \botrule
    \end{tabular}
    \label{tab:relatedmethods}
\end{table}

\section{Proposed Method}\label{sec:proposedmethod}
In this section, we introduce \gls{pafer}, a novel method to estimate the fairness of \glspl{dt} in privacy constrained manner. The following subsections dissect the proposed method, starting with \autoref{subsec:scenario}, on the assumptions and specific scenarios for which the method is built. Successively, \autoref{subsec:paferprocedure} provides a detailed description of the procedure, outlining the pseudocode and some theoretical properties. 

\subsection{Scenario}\label{subsec:scenario}
\gls{pafer} requires a specific, albeit common, scenario for its use. This subsection describes that scenario and discusses how common the scenario actually is. 

\subsubsection{Assumptions}
\gls{pafer} is a method that requires a certain setting, which comes with several assumptions. Firstly, \gls{pafer} is made for an auditing setting, in the sense that it is a method that is assumed to be used at the end of a development cycle. \gls{pafer} does not mitigate bias, it merely estimates the fairness of the rules in a \gls{dt}. Secondly, we assume that a developer has constructed a \gls{dt} that makes binary decisions on a critical task (e.g., about people). The developer may have had access to a dataset containing individuals and some task-specific features, but this dataset does not contain a full specification of sensitive attributes on an instance level. The developer (or the algorithm auditor) wants to assess the fairness of their model using \gls{sp}. We lastly assume that a legal, trusted third party exists that knows these sensitive attributes on an instance or aggregate level,\footnote{These sensitive data can be kept at the aggregate level at the legal party to minimize sensitive data leakage and to conform to privacy laws.} and is willing to share them using some safe private protocol. Based on these assumptions, the fairness of the \gls{dt} can be assessed, using the third party and \gls{pafer}. 

\subsubsection{Prevalance of Scenario}
The scenario that was described in the previous subsection can occur in the real world under varying circumstances. This subsection enumerates some assumptions and their prevalence in the real world. Firstly, it is common to see a rule-based method built for a sensitive task~\cite{navada_overview_2011, wang_bayesian_2017}. Rules are able to explain the decision process, allowing individuals that are affected by the system to receive explanations about the decision affecting them. Secondly, binary decision-making is also quite common for sensitive tasks. Prominent examples include university acceptance decision making~\cite{bickel_1975_sex}, recidivism prediction~\cite{mattu_machine_2016} and loan application evaluations~\cite{zhu_2019_study}. Moreover, multiclass decision-making problems can be rewritten as binary decision problems, as shown in \autoref{cor:binary-trees}. Thirdly, it is often the case that model developers do not have access to sensitive attributes. Simply because of regulations \cite{gdpr_2016}, or because they were not deemed necessary when gathering the data. Lastly, it is quite common that a developer worries about fairness after the construction of their model. This may be due to newly imposed regulations~\cite{aiact_2021}, due to a compliance check by an auditing body or due to newly created awareness of machine bias \cite{mattu_machine_2016}. Furthermore, when sensitive data is absent, the development of a fair rule-based system becomes difficult. There are currently no fair, interpretable, sensitive attribute agnostic classifiers, as is apparent from~\autoref{sec:relatedwork}. 

What is uncommon, however, is a third party that has the sensitive attribute data of the individuals in the dataset, albeit at an aggregate level, and is also willing to share them. As data is the new oil fueling modern machines ~\cite{szczepanski_oil_2020}, sharing data becomes more and more difficult. Since, however, fair and interpretable sensitive attribute agnostic classifiers are currently lacking, this assumption becomes necessary. This work can thus be seen as an exploration of this cooperation between the developer and the data holder, to determine the privacy risks and utility of such an exchange.

\subsection{Privacy-Aware Fairness Estimation of Rules: PAFER}\label{subsec:paferprocedure}
We propose Privacy-Aware Fairness Estimation of Rules (\gls{pafer}), a method based on \gls{dp} \cite{dwork_algorithmic_2013}, that enables the calculation of \gls{sp} for \glspl{dt} while guaranteeing privacy. \gls{pafer} sends specifically designed queries to a third party to estimate \gls{sp}. \gls{pafer} sends one query for each decision-making rule and one query for the overall composition of the sensitive attributes. The size of each (un)privileged group, along with the total number of accepted individuals from each (un)privileged group, allows us to calculate the \gls{sp}. Let $\mathcal{X}$ be the data used to train a \gls{dt}, with $x_{i}^{j}$ the $j$th feature of the $i$th individual. Let a rule be of the form $x^1 < 5 \, \land \, x^2 = True$. The query then asks for the distribution of the sensitive attributes for all individuals that have properties $x^1 < 5$ and $x^2 = True$. In \gls{pafer}, each query is a histogram query as a person cannot be both privileged and unprivileged. The query to determine the general sensitive attribute composition of all individuals can be seen as a query for an `empty' rule; a rule that applies to everyone\footnote{In logic this rule would be a tautology, a statement that is always true, e.g.\ $x^1 < 5 \lor x^1 \geq 5$. }. It can also be seen as querying the root node of a \gls{dt}.

\subsubsection{PAFER and the privacy budget}
A property of \glspl{dt} is that only one rule applies to a person. Therefore, \gls{pafer} queries each decision-making rule without having to share the privacy budget between these queries. Although we calculate a global statistic in \gls{sp}, we query each decision-making rule. This is possible due to some noise cancelling out on aggregate, and, for \glspl{dt}, because we can share the privacy budget over all decision-making rules. This intuition was also noted in~\cite{fletcher_decision_2020}. 

Because \gls{pafer} queries every individual at least once, half of the privacy budget is spent on the query to determine the general sensitive attribute composition of all individuals, and the other half is spent on the remaining queries. Still, reducing the number of queries reduces the total amount of noise. \gls{pafer} therefore prunes non-distinguishing rules. A redundant rule can be formed when the splitting criterion of the \gls{dt} improves but the split does not create a node with a different majority class. 

\subsubsection{PAFER and Statistical Parity}
The definition of \gls{sp} that \gls{pafer} calculates differs slightly from the most common, original definition  \cite{dwork_fairness_2012}, to support intersectional fairness analyses and to ensure the \gls{sp} value is in $[0, 1]$. When $A$ is a $\mathcal{K}$-ary sensitive attribute, the metric that \gls{pafer} calculates is:
\begin{equation}
    \text{SP} = \min\left(\frac{p(\hat{Y}=1|A=a)}{p(\hat{Y}=1|A=b)}\right), \, \, a,b \in \{0, 1, 2, \dots, k-1\}, a \neq b.
\end{equation}
The \gls{sp} value is always in $[0,1]$, as we arrange the fraction such that the smallest `acceptance rate' is in the numerator and the largest is in the denominator. 

\subsubsection{DP mechanisms for PAFER}
Three commonly used \gls{dp} mechanisms are apt for \gls{pafer}, namely the Laplacian mechanism, the Exponential mechanism and the Gaussian mechanism. The Laplacian mechanism is used to perform a histogram query and thus has a sensitivity of 1~\cite{dwork_algorithmic_2013}. The Exponential mechanism uses a utility function such that $u_D(r) = q(D) - |q(D) - r|$ where $r$ ranges from zero to the number of individuals that the rule applies to, and $q(D)$ is the true query answer. The sensitivity is 1 as it is based on its database argument, and this count can differ by only 1~\cite{dwork_algorithmic_2013}. The Gaussian mechanism is also used to perform a histogram query and has a sensitivity of 2, as it uses the $\Delta_2$-sensitivity. 

\subsubsection{Invalid Answer Policies}\label{subsubsec:invalidpolicies}
The Laplacian mechanism and Gaussian mechanism add noise in such a way that invalid query answers may occur. A query answer is invalid if it is negative, or if it exceeds the total number of instances in the dataset\footnote{Note that is common for a histogram query answer to exceed the number of individuals in a decision node by a certain amount. We, therefore, do not deem it as an invalid query answer.}. A policy for handling these invalid query answers must be chosen. In practice, these are mappings from invalid values to valid values. We provide several options in this subsection.

\begin{table}[htp!]
    \centering
    \caption[The proposed policy options for each type of invalid query answer. ]{The proposed policy options for each type of invalid query answer.  }\label{tab:invalidpolicies}
    \begin{tabular}{ll}
    \toprule
    Negative      & Too Large       \\ 
    \midrule
    0             & uniform         \\
    1             & total - valid   \\
    uniform       &                 \\
    total - valid &                 \\
    \botrule
    \end{tabular}
\end{table}

\autoref{tab:invalidpolicies} shows the available options for handling invalid query answers. A policy consists of a mapping chosen from the first column and a mapping chosen from the second in this table. The first column shows policies for negative query answers and the second column shows policies for query answers that exceed the number of individuals in the dataset. The `uniform' policy replaces an invalid answer with the answer if the rule would apply to the same number of individuals from each un(privileged) group. The `total - valid' policy requires that all other values in the histogram were correct and thus together allow for a calculation of the missing value by subtracting it from the total. 

\algrenewcommand{\algorithmiccomment}[1]{$\vartriangleright$ #1}
\begin{algorithm}[t]
    \caption{\gls{pafer}}\label{alg:pafer}
    \begin{algorithmic}[1]
        \Procedure{PAFER}{$\mathcal{A}, D, \varepsilon, DT, \pi, \mathcal{K}$}
        \State\Comment{$\mathcal{A}$ is a \gls{dp} mechanism that introduces noise}
        \State \Comment{$D$ is a database with $N$ instances} 
        \State \Comment{$\varepsilon$ is the privacy budget}
        \State \Comment{$DT$ is a binary Decision Tree composed of rules}
        \State \Comment{$\pi$ is a policy that transforms invalid query answers to valid query answers}
        \State \Comment{$\mathcal{K}$ is the number of sensitive groups for the sensitive attribute}
        \State $accept\_rates \gets zeros(1, \mathcal{K})$ \Comment{$accept\_rates$ is a row vector of dimension $\mathcal{K}$, initialized at 0}
        \State $total \gets \mathcal{A}(True, D, \frac{1}{2}\varepsilon)$
        \For{$q \in DT$}
            \If{$q$ is favorable}
                \State $accept\_rates$ += $\frac{\pi(\mathcal{A}(q, D, \frac{1}{2}\varepsilon))}{total}$
            \EndIf
        \EndFor 
        \State $\widehat{SP} = \frac{\min(accept\_rates)}{\max(accept\_rates)}$
        \State \textbf{return} $\widehat{SP}$
        \EndProcedure
    \end{algorithmic}
\end{algorithm}

\subsubsection{PAFER Pseudocode}\label{subsubsec:paferpseudocode}
\autoref{alg:pafer} shows the pseudocode for \gls{pafer}.

\subsubsection{Theoretical Properties of PAFER}\label{subsubsec:pafertheory}
We theoretically determine a lower and upper bound of the number of queries that \gls{pafer} requires for a $k$-ary \gls{dt} in \autoref{thm:upperlowerbound}. The lower bound is equal to two, and the upper bound is $2^{h - 1} + 1$, dependent on the height of the \gls{dt}, $h$. Note that \gls{pafer} removes redundant rules to reduce the number of rules. The larger the number of rules, the more noise is added on aggregate. 

\begin{corollary}\label{cor:binary-trees}
    Any \gls{dt} that uses non-binary splits and that classifies for a binary decision problem, can be converted to a \gls{dt} that solely uses binary splits. 
\end{corollary}

\begin{proof}
    Assume a \gls{dt} has nodes with an arbitrary number of splits  $k$, with clauses $A, B, C, \dots, K$. Converting this to a binary decision process can be achieved by chaining each clause, i.e.\ for each clause a split is created of the form $A$ or $\neg A$. The latter of the two branches is then chained to $B$ or $\neg B$, and so forth. This process is schematically shown in \autoref{fig:chaining-tree} in \autoref{sec:A1}.  Since we have proven this property for an arbitrary number of splits in a node, the property holds for any $k$-ary \gls{dt}. 
\end{proof}

\begin{theorem}\label{thm:upperlowerbound}
    The number of queries required by \gls{pafer} to estimate \gls{sp} for a binary \gls{dt} is lower bounded by 2 and upper bounded by $2^{h - 1} + 1$. 
\end{theorem}

\begin{proof}
    Assume that we have constructed a \gls{dt} for a binary classification task. By \autoref{cor:binary-trees}, the \gls{dt} can be converted to a binary tree, since it classifies for a binary classification problem. Further, let the height that this (converted) binary \gls{dt} has be $h$. To estimate \gls{sp}, for each sensitive attribute the total size is required, $|D_{A=a}|$, as well as the number of individuals from each (un)privileged group that is classified favorably by the \gls{dt}. By definition, the first quantity requires 1 histogram query. The latter quantity requires a query for each favorable decision rule in the tree. A branching node that creates one leaf node and one other branching node, adds either an unfavourable or a favourable classification rule to its \gls{dt}. The most shallow binary tree is schematically shown in \autoref{fig:lowerbound-tree} in \autoref{sec:A1}. Only 1 histogram query is required for this tree, thus the lower bound for the number of required queries for \gls{pafer} is $1 + 1 = 2$. A perfectly balanced binary tree is shown in \autoref{fig:upperbound-tree} in \autoref{sec:A1}. In this case, the number of favourable decision rules in the tree is $\frac{1}{2} 2^h = 2^{-1}2^h = 2^{h-1}$. As, by the properties of \gls{pafer}, each split that creates two leaf nodes adds both a favourable and an unfavourable classification rule to the \gls{dt}. In a perfectly balanced tree (amongst others), all nodes at $h - 1$ are such nodes. Half of the nodes at $h$ (i.e., leaf nodes) are thus favourable and half are unfavourable. This amounts to $2^{h-1}$ histogram queries. The upper bound for the number of required queries for \gls{pafer} is thus $2^{h-1} + 1$. 
\end{proof}

\section{Evaluation}\label{sec:evaluation}
This section evaluates the proposed method in the previous section, \gls{pafer}. Firstly, \autoref{subsec:experimentalsetup} describes the experimental setup, detailing the used datasets and the two experiments. Secondly, \autoref{subsec:results} displays and discusses the results of the experiments.

\subsection{Experimental Setup}\label{subsec:experimentalsetup}
This section describes the experiments that answer the research questions. The first subsection describes these datasets and details their properties. The subsections thereafter describe the experiments in order, corresponding to the research question they aim to answer. 

\subsubsection{Datasets}\label{subsubsec:datasets}
This subsection describes the datasets that are used to answer the research questions. The datasets form the test bed on which the experiments can be performed. We chose three datasets, namely Adult~\cite{kohavi_uci_2016}, COMPAS~\cite{mattu_machine_2016} and German~\cite{hoffman_uci_2013}. They are all well known in the domain of fairness for \gls{ml}, and can be considered benchmark datasets. Importantly, they vary in size and all model a binary classification problem, enabling the calculation of various fairness metrics. The datasets are publicly available and pseudonymized; every privacy concern is thus merely for the sake of argument. \autoref{tab:datasets} shows some other important characteristics of each dataset.

\begin{table}[htp!]
    \caption[Properties of the three chosen publicly available datasets.]{Properties of the three chosen publicly available datasets. } \label{tab:datasets}
    \begin{tabular}{lccccc} \toprule
        Dataset & \# Rows & \# Features & Sens. attrib. & Task\\
        \midrule
        Adult   & 48842 & 14 & \begin{tabular}[c]{@{}c@{}}race, sex, age, \\ country of origin \end{tabular} & Income $>$ \$50\,000 \\
        COMPAS  & 7214 & 53 & race, sex, age & Recidivism after 2 years \\
        German  & 1000 & 24 & \begin{tabular}[c]{@{}c@{}}race, sex, age, \\ country of origin \end{tabular} & Loan default \\ 
        \botrule     
    \end{tabular}
\end{table}

\paragraph{Pre-processing}
This paragraph describes each pre-processing step for every chosen dataset. Some pre-processing steps were taken for all datasets. In every dataset, the sensitive attributes were separated from the training set. Every sensitive attribute except age was binarized, distinguishing between privileged and unprivileged groups. The privileged individuals were White men who lived in their original country of birth, and the unprivileged individuals were those who were not male, not White or lived abroad. We now detail the pre-processing steps that are dataset-specific. 

\textit{Adult.} The Adult dataset comes with a predetermined train and test set. The same pre-processing steps were performed on each one. Rows that contained missing values were removed. The ``fnlwgt'' column, which stands for ``final weight'' was removed as it is a relic from a previously trained model and unrelated features might cause overfitting. The final number of rows was 30162 for the train set and 15060 for the test set. 

\textit{COMPAS.} The COMPAS article analyzes two datasets, one for general recidivism and one for violent recidivism~\cite{mattu_machine_2016}. Only the dataset for general recidivism was used. This is a dataset with a large number of features (53), but by following the feature selection steps from the article\footnote{\url{https://github.com/propublica/compas-analysis/blob/master/Compas\%20Analysis.ipynb}}, this number reduced to eleven, of which three are sensitive attributes. The other pre-processing step in the article is to remove cases in which the arrest date and COMPAS screening date are more than thirty days apart. The features that contain dates are then converted to just the year, rounded down. Missing values are imputed with the median value for that feature. Replacing missing values with the median value ensures that no out-of-the-ordinary values are added to the dataset. The final number of rows was 4115 for the train set and 2057 for the test set, totalling 6172 rows. 

\textit{German.} The German dataset is a nearly perfect dataset for our purposes; it contains no missing values. The gender attribute is encoded in the marital status attribute, which required separation. The final number of rows is 667 for the train set and 333 for the test set, totalling 1000 rows. 

\subsubsection{Experiment 1: Comparison of DP mechanisms for PAFER}\label{subsubsec:experimentone}
Experiment 1 was constructed such that it answered \textbf{RQ1}; what \gls{dp} mechanism is optimal for what privacy budget? The best performing shallow \gls{dt} was constructed for each dataset, using grid search and cross-validation, optimizing for balanced accuracy. The height of the \gls{dt}, the number of leaf nodes and the number of selected features were varied. The parameter space can be described as \{2, 3, 4\} $\times$ \{3, 4, 5, 6, 7, 8, 9, 10, 11, 12\} $\times$ \{sqrt, all, $\log_2$\}, constituting tuples of (height, \# leaf nodes, \# selected features). The out-of-sample \gls{sp} of each \gls{dt} is also provided in \autoref{tab:spofdts}. The experiment was repeated fifty times with this same \gls{dt}, such that the random noise, introduced by the \gls{dp} mechanisms, could be averaged. Initially, we considered the Laplacian, Exponential and Gaussian mechanisms for the comparison. However, after exploratory testing, we deemed the Gaussian mechanism to perform too poorly to be included. \autoref{tab:prelimrsq0} shows some of these preliminary results. The performance of each mechanism was measured using the \gls{aaspe}, defined as follows:
\begin{equation}\label{eq:avgabssperror}
    \mathrm{AASPE} = \sum_{i}^{\mathrm{\# \, runs}}\frac{1}{\mathrm{\# \, runs}} | SP_i - \widehat{SP_i} |,
\end{equation}
where \# runs is the number of times the experiment was repeated, $SP_i$ and $\widehat{SP_i}$ are the true and estimated $SP$ of the $i$th run, respectively. The metric was calculated out of sample, i.e., on the test set. The differences in performance were compared using an independent t-test. The privacy budget was varied such that forty equally spaced values were tested with $\varepsilon \in (0, \frac{1}{2}]$. Initial results showed that privacy budgets larger than $\frac{1}{2}$ offered very marginal improvements. \autoref{tab:prelimrsq0} shows a summary of the preliminary results for Experiment 1. Experiment 1 was performed for both ethnicity, sex and the two combined. The former two sensitive features were encoded as a binary feature, distinguishing between a privileged (white, male) and an unprivileged (non-white, non-male) group. The latter sensitive feature was encoded as a quaternary feature, distinguishing between a privileged (white-male) and an unprivileged (non-white or non-male) group.  Whenever a query answer is invalid, as described in \autoref{subsubsec:invalidpolicies}, a policy must be chosen for calculation of the \gls{sp} metric. In Experiment 1, the uniform answer approach was chosen, i.e., the size of the group was made to be proportional to the number of sensitive features and the total size. The proportion of invalid query answers, i.e., $\frac{\mathrm{\# \; invalid \; answers}}{\mathrm{\# \; total \; answers}}$, was also tracked during this experiment. This invalid value ratio provides some indication of how much noise is added to the query answers. 

\begin{table}[b!]
    \centering
    \caption[The out-of-sample Statistical Parity of each constructed Decision Tree in Experiment 1.]{The out-of-sample Statistical Parity of each constructed \gls{dt} in Experiment 1. Note that the Sex-Ethnicity attribute is encoded using four (un)privileged groups, and the others are encoded using two. }\label{tab:spofdts}
    \begin{tabular}{|c||c|c|c|}\hline
        \backslashbox{$A$}{Dataset}
        &\makebox[3em]{Adult}&\makebox[3.5em]{\small COMPAS}&\makebox[3em]{German}\\\hline\hline
        Ethnicity & 0.65 & 0.78 & 0.90  \\\hline
        Sex & 0.30 & 0.84 & 0.90  \\\hline
        Sex-Ethnicity & 0.23 & 0.72 & 0.78 \\\hline
    \end{tabular}
\end{table}

\begin{table}[b!]
    \caption[Preliminary results for Experiment 1 with larger privacy budgets.]{Preliminary results for Experiment 1 with larger privacy budgets. The Gaussian mechanism was tested with $\delta = \frac{1}{1000}$. The performance was measured using the \gls{aaspe} on the Adult dataset.}\label{tab:prelimrsq0}
    \centering
    \begin{tabular}{cccccc} \\
        \toprule
        $\varepsilon$ & Laplacian & Exponential & Gaussian & Gauss. Invalid Ratio \\
        \midrule
        0.50 & 0.02320 & 0.34350 & - & - \\
        0.55 & 0.02065 & 0.30289 & 0.32484 & 0.330 \\
        0.60 & 0.01872 & 0.25780 & 0.28916 & 0.305 \\
        0.65 & 0.01329 & 0.27566 & 0.26961 & 0.230 \\
        0.70 & 0.01026 & 0.30831 & 0.27676 & 0.250 \\
        0.75 & 0.01353 & 0.32444 & 0.26572 & 0.260 \\
        \botrule
    \end{tabular}
\end{table}

\subsubsection{Experiment 2: Comparison of different DTs for PAFER}
Experiment 2 was constructed in such a way that it answered \textbf{RQ2}; what is the effect of \gls{dt} hyperparameters on the performance of \gls{pafer}? The \texttt{minleaf} value was varied such that eighty equally spaced values were tested with $\texttt{minleaf} \in (0, \frac{1}{5}]$. In the initial results, shown in \autoref{tab:prelimrsq1}, when the \texttt{minleaf} value exceeded $\frac{1}{5}$, the same split was repeatedly chosen for each dataset. Even though \texttt{minleaf} $< \frac{1}{2}$, a risk still occurs that one numerical feature is split over and over, which hinders interpretability. Therefore, each numerical feature is categorized by binning it. The bins were established by generating five different \glspl{dt}, that used all the numerical features. An average splitting value was determined for each height across \glspl{dt}, that was kept at a maximum of seven\footnote{Based on the ``Magic Number 7", as humans can generally hold seven \textpm~two pieces of information in memory, and thus, also, seven rule clauses in memory~\cite{miller_magical_1956}.}. Averages were rounded to the nearest natural number. The privacy budget was defined such that $\varepsilon \in \{\frac{1}{20}, \frac{2}{20}, \frac{3}{20}, \frac{4}{20}, \frac{5}{20}\}$. The performance was again measured in \gls{aaspe}, as shown in \autoref{eq:avgabssperror}. The metric was measured out of sample, i.e., on the test set. The performance for each \texttt{minleaf} value was averaged over fifty potentially different \glspl{dt}. The same invalid query answer policy was chosen as in Experiment 1, replacing each invalid query answer with the uniformly distributed answer. The performance of \gls{pafer} was compared with a baseline that uniformly randomly guesses an \gls{sp} value in the interval $[0, 1)$. A one-sided t-test determined whether \gls{pafer} significantly outperformed the random baseline. 

\begin{table}[t]
    \centering
    \caption[Preliminary results for Experiment 2 with larger \texttt{minleaf} values.]{Preliminary results for Experiment 2. The performance was measured using \gls{aaspe} on the Adult dataset. The results were averaged over 25 runs. }\label{tab:prelimrsq1}
    \begin{tabular}{|l||*{5}{c|}}\hline
        \backslashbox{\texttt{minleaf}}{$\varepsilon$}
        &\makebox[3em]{$\frac{1}{20}$}&\makebox[3em]{$\frac{1}{10}$}&\makebox[3em]{$\frac{3}{20}$}
        &\makebox[3em]{$\frac{1}{5}$}&\makebox[3em]{$\frac{1}{4}$}\\\hline\hline
        $\frac{1}{5}$ & .0828 & .0532 & .0407 & .0323 & .0194 \\\hline
        $\frac{1}{4}$ & .0711 & .0325 & .0235 & .0187 & .0119 \\\hline
        $\frac{3}{10}$ & .0486 & .0282 & .0188 & .0149 & .0128 \\\hline
    \end{tabular}
\end{table}

\paragraph{Experiment 2.1: Interaction between $\varepsilon$ and \texttt{minleaf} hyperparameters}\label{par:experimenttwopointone}
The \gls{sp} metric is also popular due to its legal use in the United States, where it is used to determine compliance with the 80\%-rule \cite{eighty_rule_1979}. Thus, the \gls{uar} of \gls{pafer} was calculated for each \texttt{minleaf} value, to obtain an indication of whether \gls{pafer} was able to effectively measure this compliance. \gls{uar} is the average of class-wise recall scores. This was done by rounding each estimation down to its decimal value, thus creating `classes' that the \gls{uar} could be calculated for. To gain more intuition about the interaction between $\varepsilon$ and \texttt{minleaf} value, the following metric was calculated for each combination:
\begin{equation}\label{eq:heatmapmetric}
    \mathrm{UAR} - \mathrm{AASPE} = \sum_{c \in C} \frac{1}{|C|}\times \frac{\# \, \mathrm{true} \, c}{\# c} - \sum_{i}^{\mathrm{\# \, runs}}\frac{1}{\mathrm{\# \, runs}} | SP_i - \widehat{SP_i} | 
\end{equation}

Ideally, \gls{aaspe} is minimized and \gls{uar} is maximized, thus maximizing the metric shown in \autoref{eq:heatmapmetric}. Besides the metric, the experimental setup was identical to Experiment 2. Therefore, the same \glspl{dt} were used for this experiment, only the metrics differed. 

\subsection{Results}\label{subsec:results}
This section describes the results of the experiments and also provides an analysis of the results. Results are ordered to match the order of the experiments. 

\subsubsection{Results for Experiment 1}
\begin{figure}[t!]
    \includegraphics[width=0.99\textwidth]{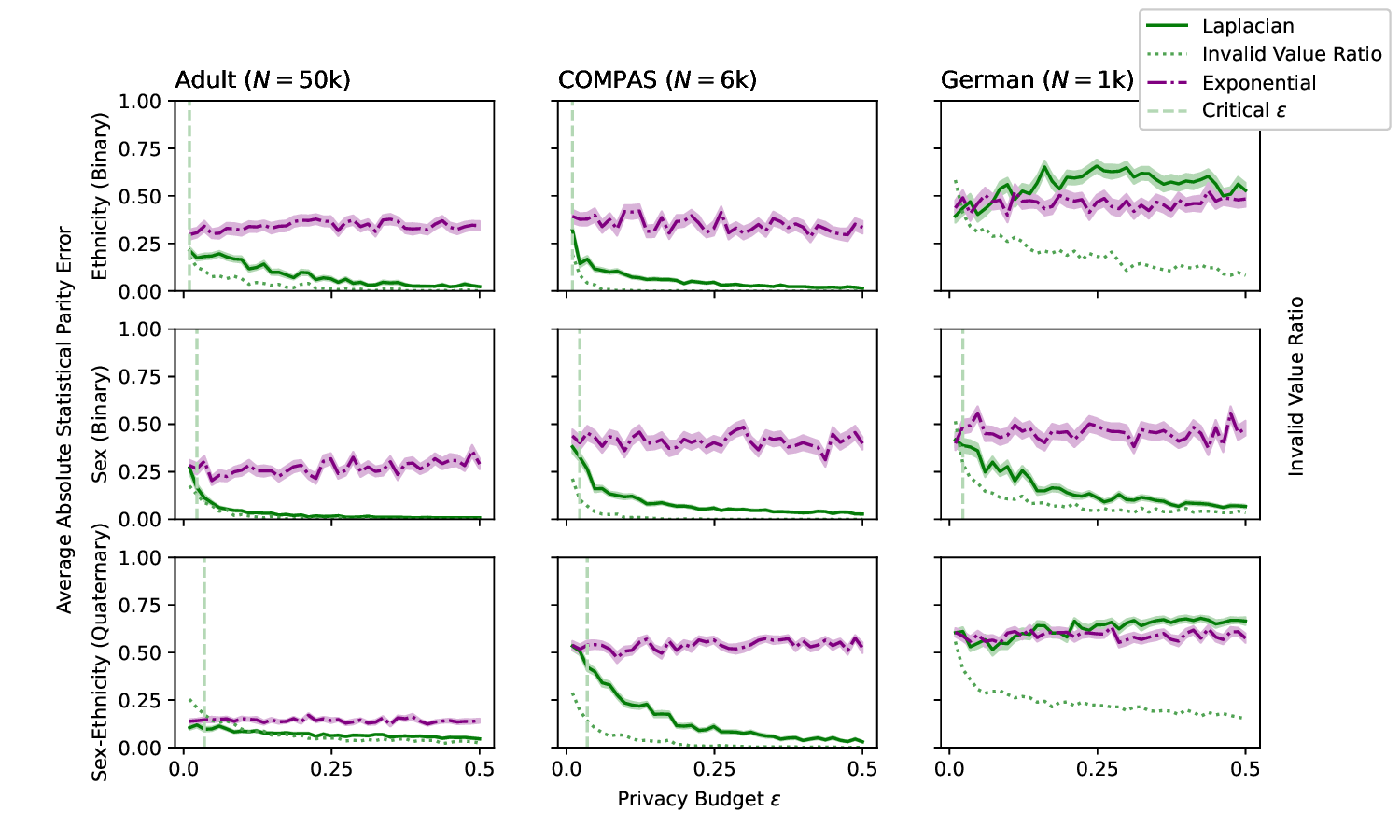}
    \caption[A comparison of the Laplacian and Exponential Differential Privacy mechanisms. ]{A comparison of the Laplacian and Exponential \gls{dp} mechanism for different privacy budgets $\varepsilon$. When indicated, from the critical $\varepsilon$ value to $\varepsilon = \frac{1}{2}$, the Laplacian mechanism performs significantly better ($p < .05$) than the Exponential mechanism. The uncertainty is pictured in a lighter color around the average. \label{fig:rsq0}}
\end{figure}
 
\autoref{fig:rsq0} answers \textbf{RQ1}; the Laplacian mechanism outperforms the Exponential mechanism on seven out of the nine analyses. The Laplacian mechanism is significantly better even at very low privacy budgets ($\varepsilon < 0.1$). The error of the mechanism generally decreases steadily, as the privacy budget increases. This is expected behavior. As the privacy budget increases, the amount of noise decreases. The Laplacian mechanism performs the best on the Adult and COMPAS datasets, because their invalid value ratio is small, especially for $\varepsilon > \frac{1}{10}$. 

The Exponential mechanism performs relatively stable across analyses, however, its performance is generally bad, with errors even reaching the maximum possible error for the German dataset. This is probably due to the design of the utility function, $u_D(r)$, which does not differentiate enough between good and bad answers. Moreover, the Exponential mechanism consistently adds even more noise because it guarantees valid query answers. The Laplacian mechanism does not give these guarantees, and thus relies less on the chosen policy, as described in \autoref{subsubsec:invalidpolicies}. The mechanism performs somewhat decently on the intersectional analysis for the Adult dataset. This is due to it being an easy prediction task, the Laplacian mechanism starts at a similarly low error. 

\autoref{fig:rsq0} shows that the invalid value ratio consistently decreases with the privacy budget. This behavior is expected, given that the amount of noise decreases as the privacy budget increases. The invalid value ratio is the largest in the intersectional analyses because then the sensitive attributes are quaternary. The difference between the invalid value ratio progression for the Adult and COMPAS datasets is small, whereas the difference between COMPAS and German is large. Thus, smaller datasets only become problematic for \gls{pafer} between 6000 and 1000 rows. Experiment 2 sheds further light on this question. 

For the two cases where the Exponential mechanism is competitive with the Laplacian mechanism, the invalid value ratio is also large. When the dataset is small, the sensitivity is relatively larger, and the chances of invalid query answers are larger. Note that the error is measured out-of-sample, so, for the German dataset, the histogram queries are performed on a dataset of size 333. This effect is also visible in the next experiment. 

\subsubsection{Results for Experiment 2}
\begin{table}[t!]
    \centering
    \caption[Results for Experiment 2 on the Adult dataset and the binary ethnicity sensitive attribute.]{Results for Experiment 2 on the Adult dataset and the binary ethnicity sensitive attribute. A * indicates that \gls{pafer} performed significantly better than the random baseline. }\label{tab:rsq1raceadult}
    \begin{tabular}{|l||*{5}{c|}}\hline
        \backslashbox{\texttt{minleaf}}{$\varepsilon$}
        &\makebox[3em]{$\frac{1}{20}$}&\makebox[3em]{$\frac{1}{10}$}&\makebox[3em]{$\frac{3}{20}$}
        &\makebox[3em]{$\frac{1}{5}$}&\makebox[3em]{$\frac{1}{4}$}\\\hline\hline
        $\frac{1}{1000}$ & $p < .001$* & $p < .001$* & $p < .001$* & $p = .001$* & $p = .039$* \\\hline
        $\frac{1}{100}$ & $p < .001$* & $p < .001$* & $p < .001$* & $p < .001$* & $p < .001$* \\\hline
        $\frac{1}{5}$ & $p < .001$* & $p < .001$* & $p < .001$* & $p < .001$* & $p < .001$* \\\hline
    \end{tabular}
\end{table}

\begin{table}[t!]
    \centering
    \caption[Results for Experiment 2 on the Adult dataset and the binary sex sensitive attribute.]{Results for Experiment 2 on the Adult dataset and the binary sex sensitive attribute. A * indicates that \gls{pafer} performed significantly better than the random baseline. A $\lozenge$ indicates that the random baseline performed significantly better than \gls{pafer}. }\label{tab:rsq1sexadult}
    \begin{tabular}{|l||*{5}{c|}}\hline
        \backslashbox{\texttt{minleaf}}{$\varepsilon$}
        &\makebox[3em]{$\frac{1}{20}$}&\makebox[3em]{$\frac{1}{10}$}&\makebox[3em]{$\frac{3}{20}$}
        &\makebox[3em]{$\frac{1}{5}$}&\makebox[3em]{$\frac{1}{4}$}\\\hline\hline
        $\frac{1}{1000}$ & $p = .999\lozenge$ & $p = .87$ & $p = .57$ & $p = .02$* & $p = .02$* \\\hline
        $\frac{1}{100}$ & $p < .001$* & $p < .001$* & $p < .001$* & $p < .001$* & $p < .001$* \\\hline
        $\frac{1}{5}$ & $p < .001$* & $p < .001$* & $p = .02$* & $p = .02$* & $p < .001$* \\\hline
    \end{tabular}
\end{table}

\begin{table}[t!]
    \centering
    \caption[Results for Experiment 2 on the Adult dataset and the quaternary sex-ethnicity sensitive attribute.]{Results for Experiment 2 on the Adult dataset and the quaternary sex-ethnicity sensitive attribute. A * indicates that \gls{pafer} performed significantly better than the random baseline. }\label{tab:rsq1sexraceadult}
    \begin{tabular}{|l||*{5}{c|}}\hline
        \backslashbox{\texttt{minleaf}}{$\varepsilon$}
        &\makebox[3em]{$\frac{1}{20}$}&\makebox[3em]{$\frac{1}{10}$}&\makebox[3em]{$\frac{3}{20}$}
        &\makebox[3em]{$\frac{1}{5}$}&\makebox[3em]{$\frac{1}{4}$}\\\hline\hline
        $\frac{1}{1000}$ & $p < .001$* & $p < .001$* & $p < .001$* & $p < .001$* & $p < .001$* \\\hline
        $\frac{1}{100}$ & $p < .001$* & $p < .001$* & $p < .001$* & $p < .001$* & $p < .001$* \\\hline
        $\frac{1}{5}$ & $p < .001$* & $p < .001$* & $p < .001$* & $p < .001$* & $p < .001$* \\\hline
    \end{tabular}
\end{table}

\begin{table}[t!]
    \centering
    \caption[Results for Experiment 2 on the COMPAS dataset and the binary ethnicity sensitive attribute.]{Results for Experiment 2 on the COMPAS dataset and the binary ethnicity sensitive attribute. A * indicates that \gls{pafer} performed significantly better than the random baseline. }\label{tab:rsq1racecompas}
    \begin{tabular}{|l||*{5}{c|}}\hline
        \backslashbox{\texttt{minleaf}}{$\varepsilon$}
        &\makebox[3em]{$\frac{1}{20}$}&\makebox[3em]{$\frac{1}{10}$}&\makebox[3em]{$\frac{3}{20}$}
        &\makebox[3em]{$\frac{1}{5}$}&\makebox[3em]{$\frac{1}{4}$}\\\hline\hline
        $\frac{1}{1000}$ & $p < .001$* & $p < .001$* & $p < .001$* & $p < .001$* & $p < .001$* \\\hline
        $\frac{1}{100}$ & $p < .001$* & $p < .001$* & $p < .001$* & $p < .001$* & $p < .001$* \\\hline
        $\frac{1}{5}$ & $p < .001$* & $p < .001$* & $p < .001$* & $p < .001$* & $p < .001$* \\\hline
    \end{tabular}
\end{table}

\begin{table}[t!]
    \centering
    \caption[Results for Experiment 2 on the COMPAS dataset and the binary sex sensitive attribute.]{Results for Experiment 2 on the COMPAS dataset and the binary sex sensitive attribute. A * indicates that \gls{pafer} performed significantly better than the random baseline.}\label{tab:rsq1sexcompas}
    \begin{tabular}{|l||*{5}{c|}}\hline
        \backslashbox{\texttt{minleaf}}{$\varepsilon$}
        &\makebox[3em]{$\frac{1}{20}$}&\makebox[3em]{$\frac{1}{10}$}&\makebox[3em]{$\frac{3}{20}$}
        &\makebox[3em]{$\frac{1}{5}$}&\makebox[3em]{$\frac{1}{4}$}\\\hline\hline
        $\frac{1}{1000}$ & $p = .94$ & $p = .46$ & $p = .27$ & $p = .015$* & $p < .001$* \\\hline
        $\frac{1}{100}$ & $p < .001$* & $p < .001$* & $p < .001$* & $p < .001$* & $p < .001$* \\\hline
        $\frac{1}{5}$ & $p < .001$* & $p < .001$* & $p < .001$* & $p < .001$* & $p < .001$* \\\hline
    \end{tabular}
\end{table}

\begin{table}[t!]
    \centering
    \caption[Results for Experiment 2 on the COMPAS dataset and the quaternary sex-ethnicity sensitive attribute.]{Results for Experiment 2 on the COMPAS dataset and the quaternary sex-ethnicity sensitive attribute. A * indicates that \gls{pafer} performed significantly better than the random baseline. A $\lozenge$ indicates that the random baseline performed significantly better than \gls{pafer}. }\label{tab:rsq1sexracecompas}
    \begin{tabular}{|l||*{5}{c|}}\hline
        \backslashbox{\texttt{minleaf}}{$\varepsilon$}
        &\makebox[3em]{$\frac{1}{20}$}&\makebox[3em]{$\frac{1}{10}$}&\makebox[3em]{$\frac{3}{20}$}
        &\makebox[3em]{$\frac{1}{5}$}&\makebox[3em]{$\frac{1}{4}$}\\\hline\hline
        $\frac{1}{1000}$ & $p = 1\lozenge$ & $p = 1\lozenge$ & $p = 1\lozenge$ & $p = 1\lozenge$ & $p = .98\lozenge$ \\\hline
        $\frac{1}{100}$ & $p = .38$ & $p < .001$* & $p < .001$* & $p < .001$* & $p < .001$* \\\hline
        $\frac{1}{5}$ & $p = 0.99\lozenge$ & $p < .001$* & $p < .001$* & $p < .001$* & $p < .001$* \\\hline
    \end{tabular}
\end{table}

\autoref{tab:rsq1raceadult} through \autoref{tab:rsq1sexracecompas} show the results for Experiment 2. The tables clearly show that \gls{pafer} generally significantly outperforms the random baseline. For small privacy budgets ($\varepsilon \leq \frac{1}{10}$) and small \texttt{minleaf} values (\texttt{minleaf} = $\frac{1}{1000}$), \gls{pafer} does not strictly perform better, for instance in \autoref{tab:rsq1sexcompas}. \gls{pafer} is even significantly outperformed by the random baseline in some cases, such as in \autoref{tab:rsq1sexadult} and \autoref{tab:rsq1sexracecompas}, for similarly small values of $\varepsilon$ and \texttt{minleaf}. \gls{pafer} thus performs poorly with a small privacy budget, but also on less interpretable \glspl{dt}. When the \texttt{minleaf} value of a \gls{dt} is small, it generally has more branches and branches are longer, as it takes more splits to reach the desired \texttt{minleaf} size. Both of these factors worsen the interpretability of a \gls{dt} \cite{barredo_arrieta_explainable_2020}. 

Other factors negatively impacting the performance of \gls{pafer} are a small dataset size and the number of (un)privileged groups. Therefore, the results for the German dataset are omitted, as \gls{pafer} is entirely outperformed by the random baseline. This also occurs in \autoref{tab:rsq1sexracecompas}, for all $\varepsilon$ and \texttt{minleaf} = $\frac{1}{1000}$. This is due to the smaller leaf nodes, but also due to the smaller dataset (N = 6000), and the quaternary sex-ethnicity sensitive attribute. This reduces the queried quantities even further, resulting in worse performance for \gls{pafer}. Then, the (un)privileged group sizes are closer to zero per rule, which increases the probability of invalid query answers. \gls{pafer}'s worse performance on smaller datasets, and less interpretable \glspl{dt} is a clear limitation of the method. 

For the sake of succinctness, the results and respective plots for Experiment 2.1 are given in~\autoref{secA2}. This final experiment also replicates some of the results of Experiment 1 and Experiment 2. The middle plot in \autoref{fig:rsq1gridraceadult} through \autoref{fig:rsq1gridsexracecompas} shows that \gls{pafer} with the Laplacian mechanism performs better for larger privacy budgets. These plots also show the previously mentioned trade-off between interpretability and performance of \gls{pafer}; the method performs worse for smaller \texttt{minleaf} values. Lastly, the performance is generally lower for the COMPAS dataset, which holds fewer instances.

\section{Conclusion \& Future Work}\label{sec:conclusionfutwork}
This section concludes the work with answers to the research questions in \autoref{subsec:rqsanswers}, summarizes the entire work in \autoref{subsec:summary}, and provides suggestions for future work in \autoref{subsec:futwork}. 

\subsection{Answers to the Research Questions}\label{subsec:rqsanswers}
This section will answer the research questions (RQs) and research subquestions (RSQs), as posed in \autoref{subsec:rqs}. 

\textbf{RQ1} What is the optimal privacy mechanism that preserves privacy and minimizes average Statistical Parity error? \\ 
The optimal \gls{dp} mechanism in Experiment 1 was the Laplacian mechanism, as shown in \autoref{fig:rsq0}. It performed optimally, in the sense that it achieved a low \gls{aaspe} at small privacy budgets. This varied from 0.05 error at $\varepsilon = 0.1$, to an error of 0.1 at $\varepsilon = 0.25$. The preliminary results showed that the Gaussian mechanism was also far from optimal, even for large privacy budgets (\autoref{tab:prelimrsq0}). 

\textbf{RSQ1.1} Is there a statistically significant mean difference in Absolute Statistical Parity error between the Laplacian mechanism and the Exponential mechanism? \\ 
Yes, the Laplacian mechanism significantly outperformed the Exponential mechanism at very low privacy budgets, on seven out of the nine performed analyses (\autoref{fig:rsq0}). The Gaussian mechanism proved also to be of no match for the Laplacian mechanism, even at large privacy budgets (\autoref{tab:prelimrsq0}). 

\textbf{RQ2} Is there a statistically significant difference between the Statistical Parity errors of \gls{pafer} compared to other benchmarks for varying Decision Tree hyperparameter values? \\
Yes, for nearly all trials in Experiment 2, there was a significant difference in error between \gls{pafer} and the random baseline. 

\textbf{RSQ2.1} At what fractional \texttt{minleaf} value is \gls{pafer} significantly better at estimating Statistical Parity than a random baseline? \\
The answer depends on the sensitive attribute that is analyzed and the dataset. In Experiment 2, for the Adult dataset, a \texttt fractional {minleaf} value of $\frac{1}{100}$ ensured that \gls{pafer} significantly outperformed the random baseline, (\autoref{tab:rsq1sexraceadult}). For the COMPAS dataset and intersectional analysis, a privacy budget of $\varepsilon = \frac{1}{20}$ was not enough to statistically prove that \gls{pafer} outperformed the random baseline (\autoref{tab:rsq1sexracecompas}).  

\subsection{Summary}\label{subsec:summary}
This work has shed light on the trade-offs between fairness, privacy and interpretability, by introducing a novel, privacy-aware fairness estimation method called \gls{pafer}. There is a natural tension between the estimation of fairness and privacy, given that sensitive attributes are required to calculate fairness. This applies also to interpretable, rule-based methods. The proposed method, \gls{pafer}, alleviates some of this tension. \\ 
\gls{pafer} should be applied on a \gls{dt} in a binary classification setting, at the end of a development cycle. 
\gls{pafer} guarantees privacy using mechanisms from \gls{dp}, allowing it to measure \gls{sp} for \glspl{dt}. \\
We showed that the minimum number of required queries for \gls{pafer} is 2. We also showed that the maximum number of queries depends on the height of the \gls{dt} via $2^{h-1} + 1$, where $h$ is the height. \\ 
In our experimental comparison of several \gls{dp} mechanisms, \gls{pafer} showed to be capable of accurately estimating \gls{sp} for low privacy budgets ($\varepsilon = \frac{1}{10}$) when used with the Laplacian mechanism. This confirms that the calculation of \gls{sp} for \glspl{dt} while respecting privacy is possible using \gls{pafer}. \\
Experiment 2 showed that the smaller the leaf nodes of the \gls{dt} are, the worse the performance is. \gls{pafer} thus performs better for more interpretable \glspl{dt}; as the smaller the \texttt{minleaf} value is, the less interpretable a \gls{dt} is. \\
Future work can look into other types of \gls{dp} mechanisms to use with \gls{pafer}, and other types of fairness metrics, e.g.\ \gls{eqodds}. 

\subsection{Limitations \& Future Work}\label{subsec:futwork}
This section describes some avenues that could be further explored regarding \gls{pafer}, with an eye on the limitations that became apparent from the experimental results. We suggest an extension of \gls{pafer} that can adopt two other new fairness metrics in \autoref{subsubsec:futworkmetrics} and suggest examining the different parameters of the \gls{pafer} algorithm in \autoref{subsubsec:futworkparams}. 

\subsubsection{Other fairness metrics}\label{subsubsec:futworkmetrics}
The most obvious research avenue for \gls{pafer} is the extension to support other fairness metrics. \gls{sp} is a popular, but simple metric that is not correct in every scenario. We thus propose two other group fairness metrics that are suitable for \gls{pafer}. However, with the abundance of fairness metrics, multiple other suitable metrics are bound to exist. 

The \gls{eqodds} metric compares the acceptance rates across (un)privileged groups and dataset labels. In our scenario (\autoref{subsec:scenario}), we assume to know the dataset labels, as this is required for the construction of a \gls{dt}. Therefore, by querying the sensitive attribute distributions for favorably classifying rules, only for those individuals for which $Y = y$, \gls{pafer} can calculate \gls{eqodds}. Since these groups are mutually exclusive, $\varepsilon$ does not have to be shared. Since \gls{eqopp} is a variant of \gls{eqodds}, this can naturally also be measured using this approach. A downside is that the number of queries is multiplied by a factor of two, which hinders performance. However, this is not much of an overhead because it is only a constant factor. 

\subsubsection{Other input parameters}\label{subsubsec:futworkparams}
Examining the input parameters of the \gls{pafer} estimation algorithm in \autoref{alg:pafer}, two clear candidates for further research become visible. These are the \gls{dp} mechanism, $\mathcal{A}$ and the model that is audited, $DT$. \hyperref[par:futworkdpmech]{Paragraph~\ref*{par:futworkdpmech}} and \hyperref[par:futworkdpmech]{paragraph~\ref*{par:futworkmodel}} discuss these options. 

\paragraph{The Differential Privacy mechanism}\label{par:futworkdpmech}
The performance of other \gls{dp} mechanisms can be experimentally compared to the currently examined mechanisms, using the experimental setup of Experiment 1. Experiment 2 shows that there is still room for improvement, as a random guessing baseline significantly outperforms the Laplacian mechanism on multiple occasions. 

The work of Hamman et al.\ in \cite{hamman_can_2022} shows promising results for a simple \gls{sp} query. They use a \gls{dp} mechanism based on smooth sensitivity \cite{nissim_smooth_2007}; a sensitivity that adds data-specific noise to guarantee \gls{dp}. If this \gls{dp} mechanism could be adopted for histogram queries, \gls{pafer} might improve in accuracy. Currently, \gls{pafer} improves poorly on less interpretable \glspl{dt}. An improvement in accuracy might also enable \gls{pafer} to audit less interpretable \glspl{dt}. 


\paragraph{The audited model}\label{par:futworkmodel}
\gls{pafer}, as the name suggests, is currently only suited for rule-based systems, and in particular \glspl{dt}. Further research could look into the applicability of \gls{pafer} for other rule-based systems, such as fuzzy-logic rule systems \cite{mendel_uncertain_2017}, rule lists \cite{angelino_learning_2018} and association rule data mining \cite{yazgana2016literature}. The main point of attention is the distribution of the privacy budget. For \glspl{dt}, only one rule applies to each person, so \gls{pafer} can query all rules. For other rule-based methods, this might not be the case. 

Aytekin made the connection between Neural Networks and \glspl{dt} explicit, showing that for any activation function, a Neural Network can be written as a \gls{dt}~\cite{aytekin_neural_2022}. Applying \gls{pafer} to extracted \glspl{dt} from Neural Networks could also be a future research direction. However, the Neural Network must have a low number of parameters, or else the associated \gls{dt} would be very tall. \glspl{dt} with a tall height work worse with \gls{pafer}, so the applicability is limited.

\backmatter


\section*{Declarations}


\begin{itemize}
\item Funding: The research leading to this article was conducted during an internship at the Dutch Central Government Audit Service (ADR) as part of the Utrecht University MSc thesis study of the first author.
\item Conflict of interest/Competing interests: Authors declare no conflict of interest.
\item Ethics approval: Not applicable.
\item Consent to participate: Not applicable.
\item Consent for publication: Authors received written consent from the ADR for the publication of this work.
\item Availability of data and materials: The datasets, Adult \cite{kohavi_uci_2016}, COMPAS \cite{mattu_machine_2016} and German \cite{hoffman_uci_2013},  used in the study are publicly available. 
\item Code availability: Codes of PAFER will be made publicly available upon acceptance of the paper. 
\item Authors' contributions: The research is conducted by FvdS and supervised by FV and HK. The proposed method is conceptualized by FvdS and matured in consultation with the other authors. Experiments are conducted and discussed by all three authors. The manuscript is written by FvdS, then reviewed and revised by FV and HK. All authors have read and approved the final version.
\end{itemize}

\printglossary


\begin{appendices}

\section{Figures Illustrating PAFER's Theoretical Properties}\label{sec:A1}

\setcounter{figure}{0} 
\renewcommand{\thefigure}{A\arabic{figure}}
\renewcommand{\theHfigure}{A\arabic{figure}}

\begin{figure}
    \centering
    \includegraphics{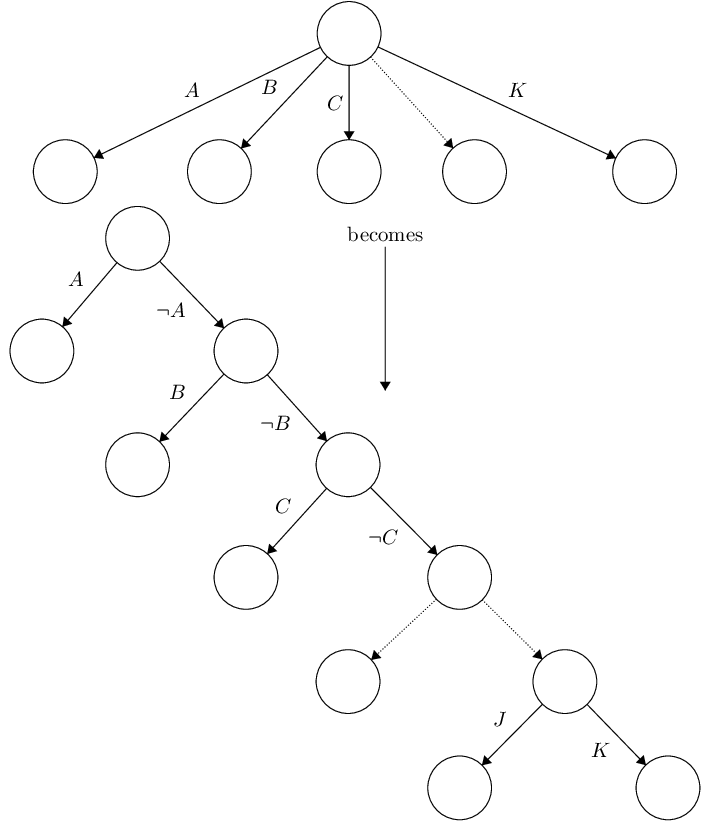}
    \caption[A schematic display of the conversion from a non-binary tree to a binary tree. ]{A schematic display of the process by which a binary tree that has non-binary splits can be converted into a binary tree for a binary decision process. The dotted lines $\iddots$, denote that the pattern of the \gls{dt} can be repeated an arbitrary number of times. \label{fig:chaining-tree}}
\end{figure}

\begin{figure}
    \centering
    \includegraphics{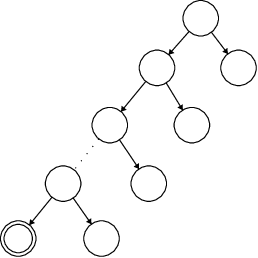}
    \caption[The smallest number of favorable decision rules in a decision tree for a binary classification problem.]{The smallest number of favorable decision rules in a decision tree for a binary classification problem. The leaf node with an inner circle denotes a leaf node in which the majority of the individuals are classified favorably in the training set. The dotted line, $\iddots$, denotes that the pattern can go on indefinitely. \label{fig:lowerbound-tree}}
\end{figure}

\begin{figure}
    \centering
    \includegraphics{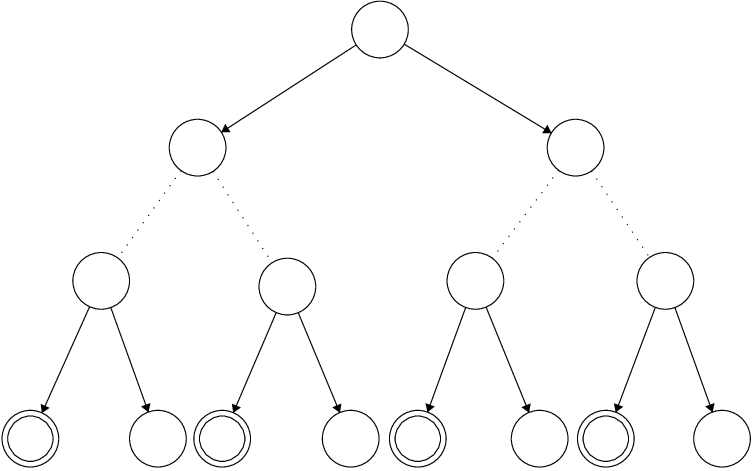}
        \caption[The largest number of favorable decision rules in a decision tree for a binary classification problem.]{The largest number of favorable decision rules in a decision tree for a binary classification problem. The leaf nodes with an inner circle denote a leaf node in which the majority of the individuals are classified favorably in the training set. The dotted lines, $\iddots$, denote that the pattern can go on indefinitely. \label{fig:upperbound-tree}}
\end{figure}

\section{Results for Experiment 2.1}\label{secA2}
\begin{figure}[htp!]
    \centering
    \includegraphics[width=0.8\textwidth]{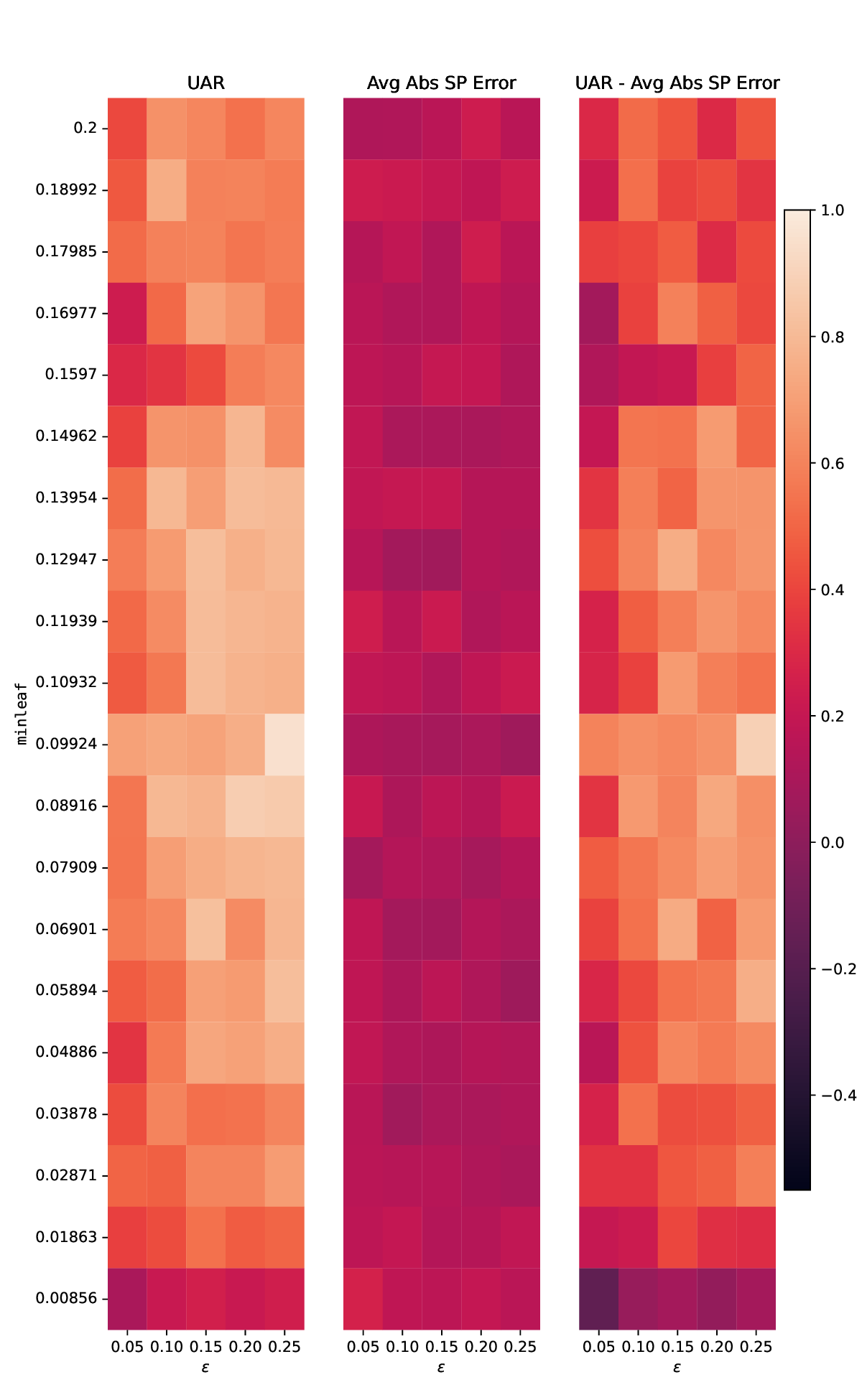}
    \caption{The hyperparameter space for the Adult dataset and the binary ethnicity attribute. }\label{fig:rsq1gridraceadult}
\end{figure}

\begin{figure}[htp!]
    \centering
    \includegraphics[width=0.8\textwidth]{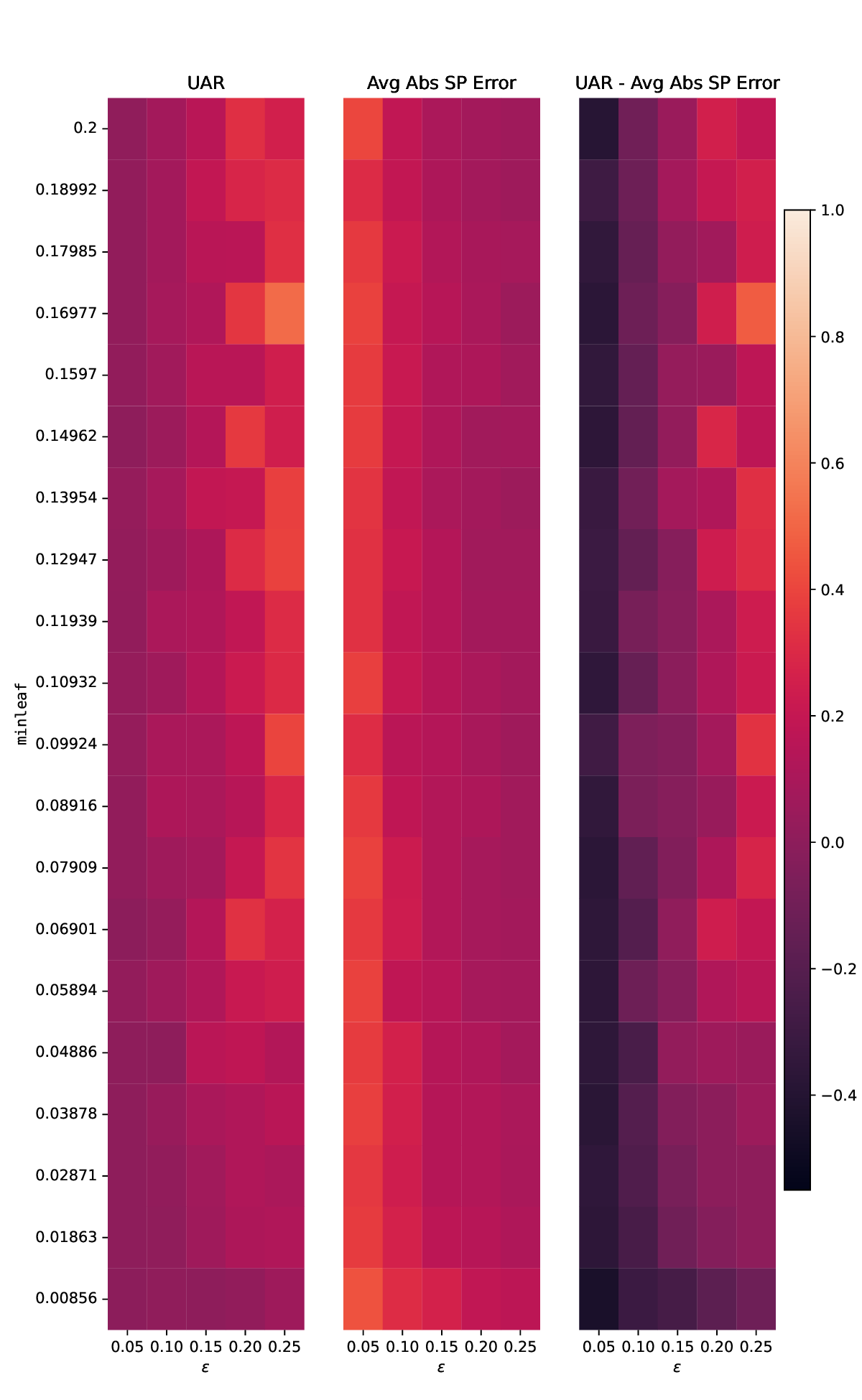}
    \caption{The hyperparameter space for the Adult dataset and the binary sex attribute. }\label{fig:rsq1gridsexadult}
\end{figure}

\begin{figure}[htp!]
    \centering
    \includegraphics[width=0.8\textwidth]{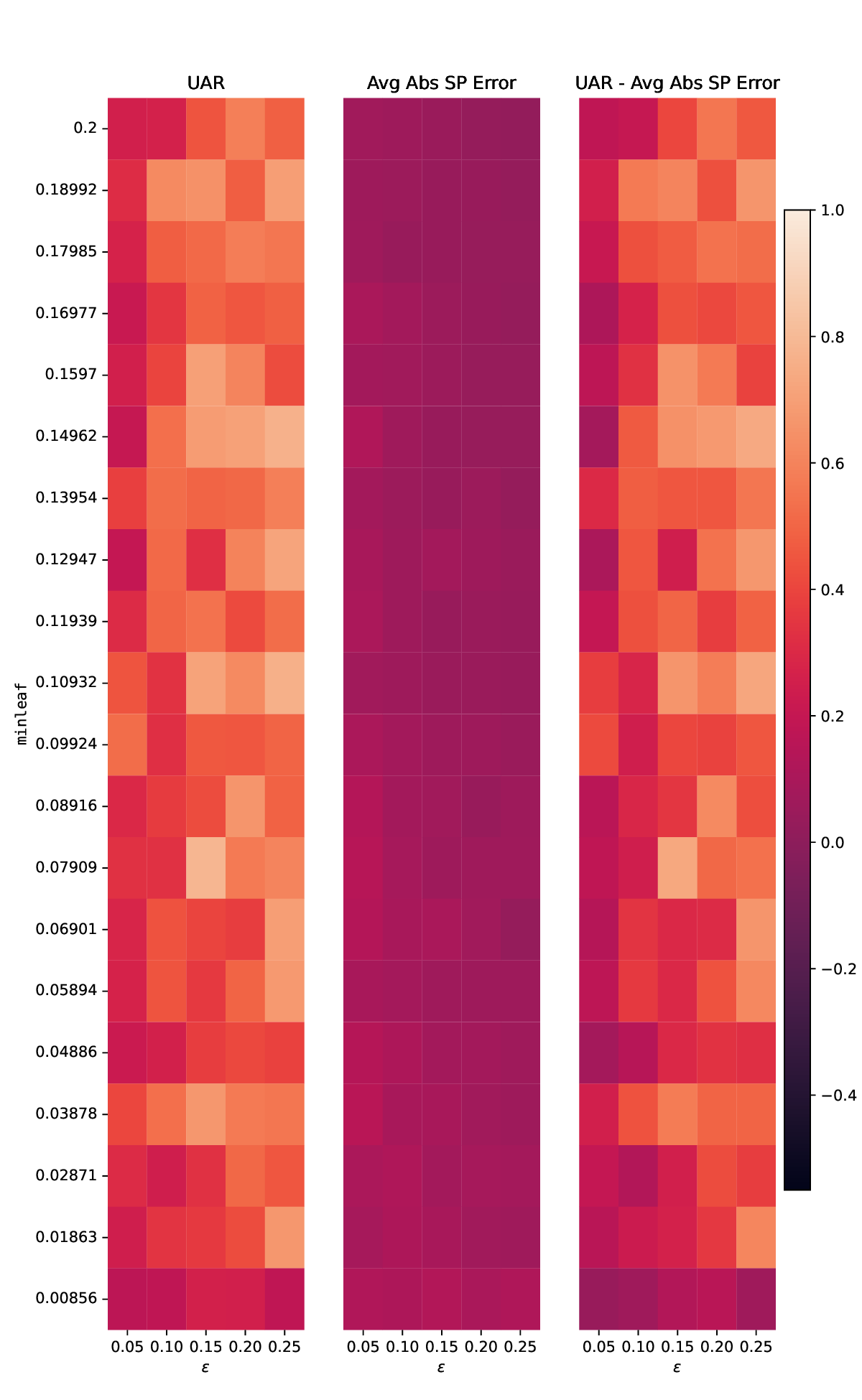}
    \caption{The hyperparameter space for the Adult dataset and the quaternary sex-ethnicity attribute. }\label{fig:rsq1gridsexraceadult}
\end{figure}

\begin{figure}[htp!]
    \centering
    \includegraphics[width=0.8\textwidth]{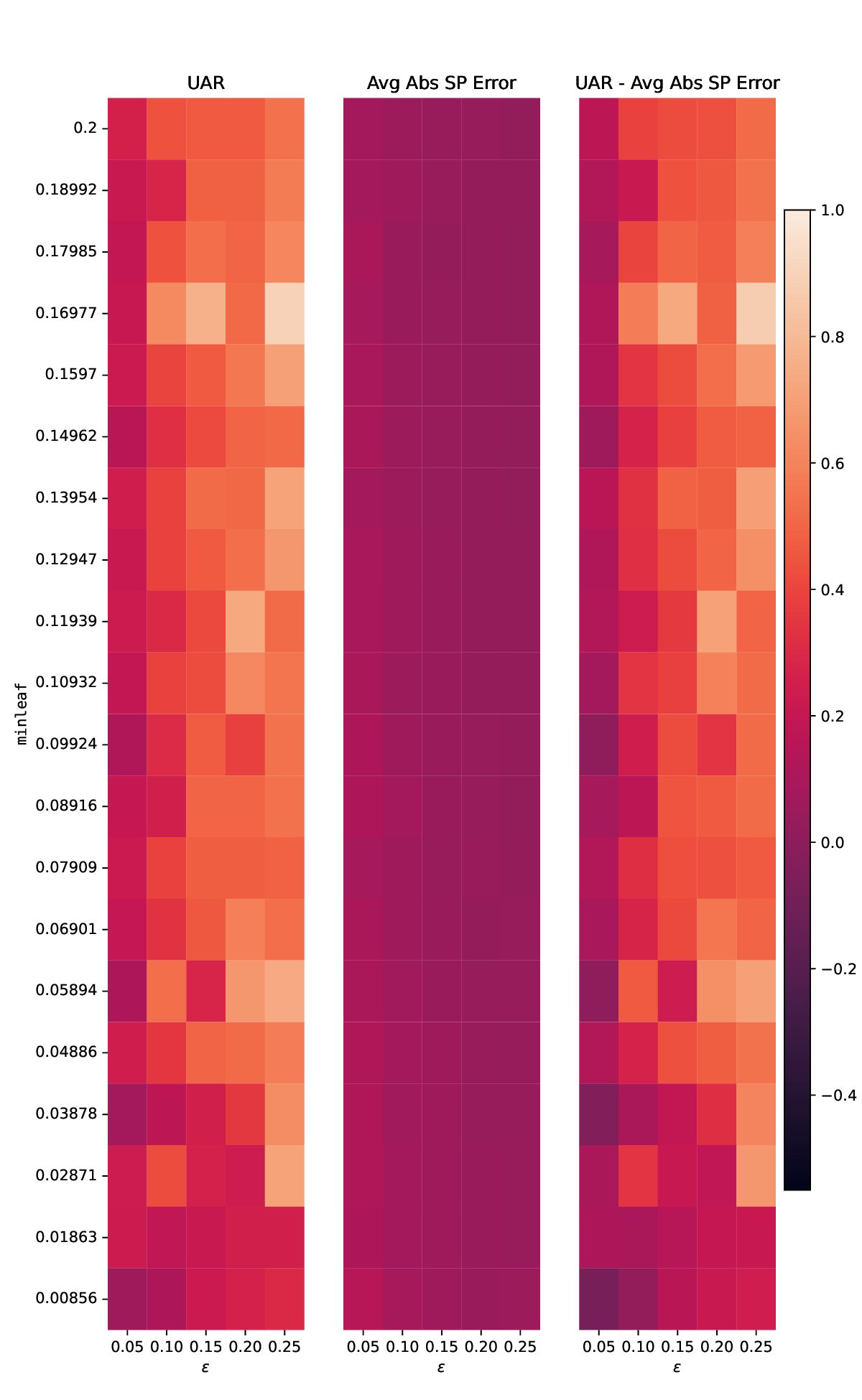}
    \caption{The hyperparameter space for the COMPAS dataset and the binary ethnicity attribute. }\label{fig:rsq1gridracecompas}
\end{figure}

\begin{figure}[htp!]
    \centering
    \includegraphics[width=0.8\textwidth]{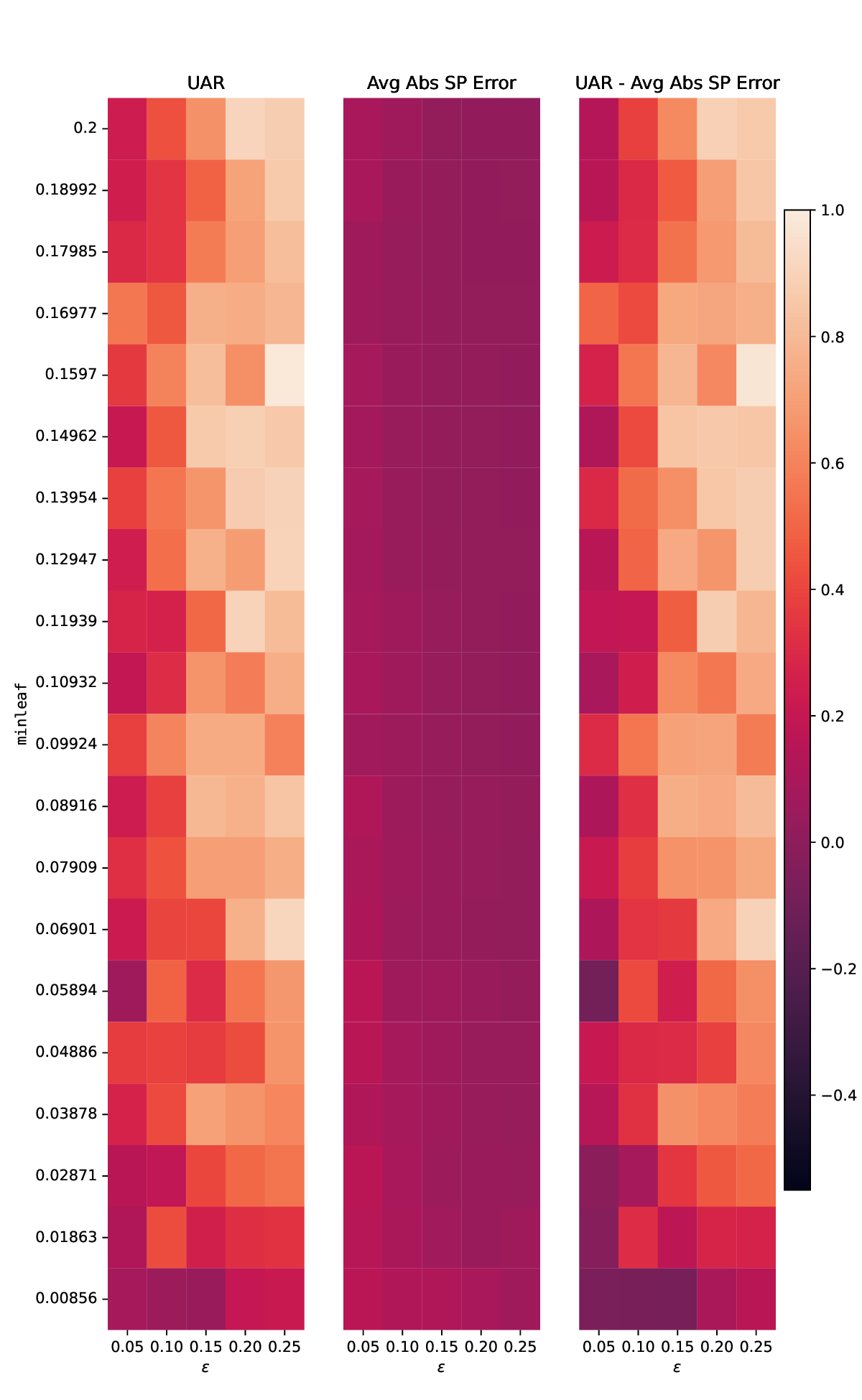}
    \caption{The hyperparameter space for the COMPAS dataset and the binary sex attribute.}\label{fig:rsq1gridsexcompas}
\end{figure}

\begin{figure}[htp!]
    \centering
    \includegraphics[width=0.8\textwidth]{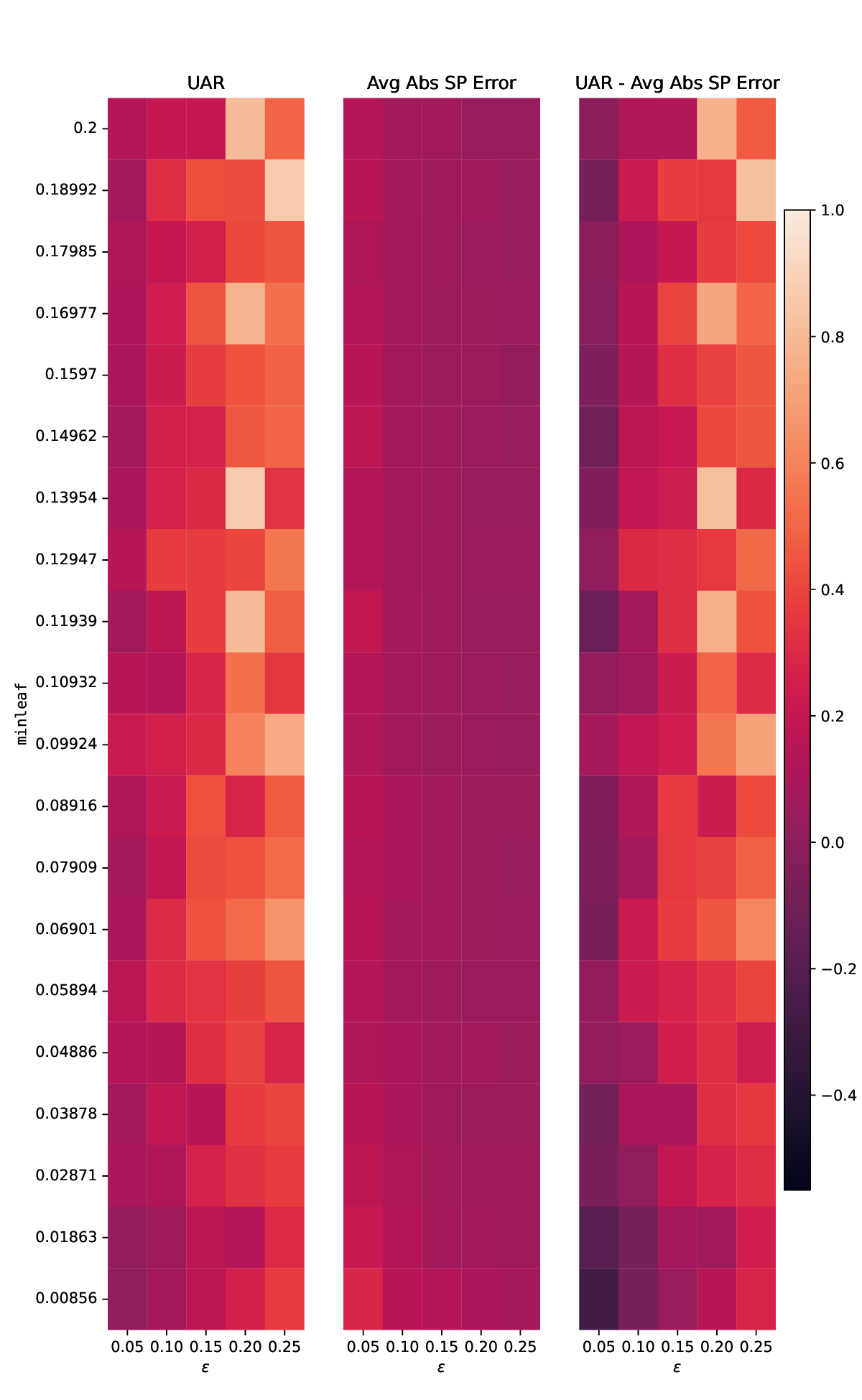}
    \caption{The hyperparameter space for the COMPAS dataset and the quaternary sex-ethnicity attribute.}\label{fig:rsq1gridsexracecompas}
\end{figure}

\autoref{fig:rsq1gridraceadult} through \autoref{fig:rsq1gridsexracecompas} show the results for Experiment 2.1. Experiment 2.1 shows that \gls{pafer} is unreliable in its ability to predict adherence to the 80\%-rule. For some datasets and sensitive attributes, \gls{pafer} performs quite well, e.g.\ reaching around 90\% \gls{uar}, as shown in \autoref{fig:rsq1gridsexcompas} and \autoref{fig:rsq1gridraceadult}. For other datasets and sensitive attributes, \gls{pafer} performs rather poorly, reaching no higher than 50\% \gls{uar} on the Adult dataset with the binary sex attribute, as shown in \autoref{fig:rsq1gridsexadult}.

Nonetheless, a pattern emerges from \autoref{fig:rsq1gridraceadult} through \autoref{fig:rsq1gridsexracecompas} regarding the \gls{uar} - \gls{aaspe}. Of course, \gls{pafer} performs better for privacy budgets larger than $\frac{3}{20}$. However, \gls{pafer} also performs better for certain \texttt{minleaf} values. The `hotspot' differs between the Adult and COMPAS dataset, \texttt{minleaf} = $\frac{1}{10}$ and \texttt{minleaf} = $\frac{3}{20}$, respectively, but the range seems to be from $\frac{7}{100}$ to $\frac{1}{5}$. The ideal scenario for \gls{pafer} thus seems to be when a privacy budget of at least $\varepsilon = \frac{3}{20}$ is available, and the examined \gls{dt} has leaf nodes with a fractional \texttt{minleaf} value of at least $\frac{7}{100}$. 

\end{appendices}



\end{document}